\documentclass[]{memtensor}

\usepackage{amsfonts}
\usepackage{amsmath}

\usepackage[dvipsnames]{xcolor}
\usepackage{tikz}
\usepackage{float}
\usepackage{fontawesome7}
\usetikzlibrary{
    shapes, 
    shapes.multipart, 
    arrows.meta, 
    positioning, 
    fit, 
    backgrounds, 
    calc,
    decorations.pathreplacing,
    shadings
}

\usepackage{lmodern}
\usepackage{graphicx}
\usepackage{array}      
\usepackage{graphicx}   
\usepackage{amssymb}
\usepackage[table]{xcolor} 
\usepackage{multirow} 
\usepackage{makecell} 
\usepackage{booktabs} 
\usepackage[dvipsnames]{xcolor} 
\usepackage{pifont}             
\usepackage{arydshln} 
\usepackage{wrapfig,lipsum}
\usepackage{enumitem}

\definecolor{headergray}{gray}{0.92} 
\definecolor{myblue}{HTML}{1F78B4}
\definecolor{mypink}{HTML}{FB9A99}
\newcommand{\cmark}{\textcolor{ForestGreen}{\ding{51}}} 
\newcommand{\xmark}{\textcolor{Maroon}{\ding{55}}}      
\definecolor{upcolor}{HTML}{228B22} 
\definecolor{downcolor}{HTML}{CC0000} 

\newcommand{\up}[1]{\textcolor{upcolor}{\scriptsize{(+#1)}}}
\newcommand{\down}[1]{\textcolor{downcolor}{\scriptsize{(-#1)}}}
\newcommand{\vhead}[1]{\rotatebox{45}{#1\hspace{2pt}}}
\newcommand{\phead}[1]{\rotatebox{0}{#1\hspace{2pt}}}
\title{CL-VISTA: Benchmarking Continual Learning in Video Large Language Models}

\author{Haiyang Guo$^{1,2}$}
\author{Yichen Shi$^{1}$}
\author{Fei Zhu$^{3}$}
\author{Wenzhuo Liu$^{2,4}$}
\author{Hongbo Zhao$^{2, 4}$}
\author{Fanhu Zeng$^{2, 4}$}
\author{Shijie Ma$^{2,4}$}
\author{Da-Han Wang$^{5}$}
\author{Xu-Yao Zhang${^{1,2,4\text{ \faIcon[regular]{envelope}}}}$}

\affiliation{$^1$School of Advanced Interdisciplinary Sciences, University of Chinese Academy of Sciences}
\affiliation{$^2$MAIS, Institute of Automation, Chinese Academy of Sciences}
\affiliation{$^3$Centre for Artificial Intelligence and Robotics, Hong Kong Institute of Science \& Innovation, CAS}
\affiliation{$^4$School of Artificial Intelligence, University of Chinese Academy of Sciences}
\affiliation{$^5$FKLPRIU, School of Computer and Information Engineering, Xiamen University of Technology}

\contribution[\text{\faIcon[regular]{envelope}}]{Corresponding Author}
\checkdata[\faIcon{github} Github]{\url{https://github.com/Ghy0501/MCITlib}}
\checkdata[\faIcon{database} Huggingface]{\url{https://huggingface.co/datasets/MLLM-CL/CL-VISTA}}
\checkdata[\faIcon{robot} ModelScope]{\url{https://www.modelscope.cn/datasets/MLLM-CL/CL-VISTA}}

\abstract{
Video Large Language Models (Video-LLMs) require continual learning to adapt to non-stationary real-world data. However, existing benchmarks fall short of evaluating modern foundation models: many still rely on models without large-scale pre-training, and prevailing benchmarks typically partition a single dataset into sub-tasks, resulting in high task redundancy and negligible forgetting on pre-trained Video-LLMs. To address these limitations, we propose CL-VISTA, a benchmark tailored for continual video understanding of Video-LLMs. By curating 8 diverse tasks spanning perception, understanding, and reasoning, CL-VISTA induces substantial distribution shifts that effectively expose catastrophic forgetting. To systematically assess CL methods, we establish a comprehensive evaluation framework comprising 6 distinct protocols across 3 critical dimensions: performance, computational efficiency, and memory footprint. Notably, the performance dimension incorporates a general video understanding assessment to assess whether CL methods genuinely enhance foundational intelligence or merely induce task-specific overfitting. Extensive benchmarking of 10 mainstream CL methods reveals a fundamental trade-off: no single approach achieves universal superiority across all dimensions. Methods that successfully mitigate catastrophic forgetting tend to compromise generalization or incur prohibitive computational and memory overheads. We hope CL-VISTA provides critical insights for advancing continual learning in multimodal foundation models.\\}

\begin{document}

\maketitle
\section{Introduction}
\label{sec:intro}

Recent advances in multimodal foundation models, particularly Video Large Language Models (Video-LLMs), which combine vision encoders with large language models, have demonstrated remarkable capabilities in video understanding and reasoning\cite{bai2025qwen3, zhang2025videollama, zhang2024llava, feng2025video}. However, most existing models are trained on static, offline datasets, creating a significant gap when faced with the non-stationary data distributions and evolving scenarios inherent in real-world deployments\cite{parmar2023open}. This discrepancy necessitates the development of Video-LLMs capable of Continual Learning (CL)\cite{guo2025continual, shi2025continual, wang2024comprehensive}, which enables models to incrementally acquire new knowledge while mitigating catastrophic forgetting\cite{french1999catastrophic, kirkpatrick2017overcoming} to ensure sustained performance across long-term task sequences.

\begin{figure}[tb]
  \centering
  \includegraphics[width=\textwidth]{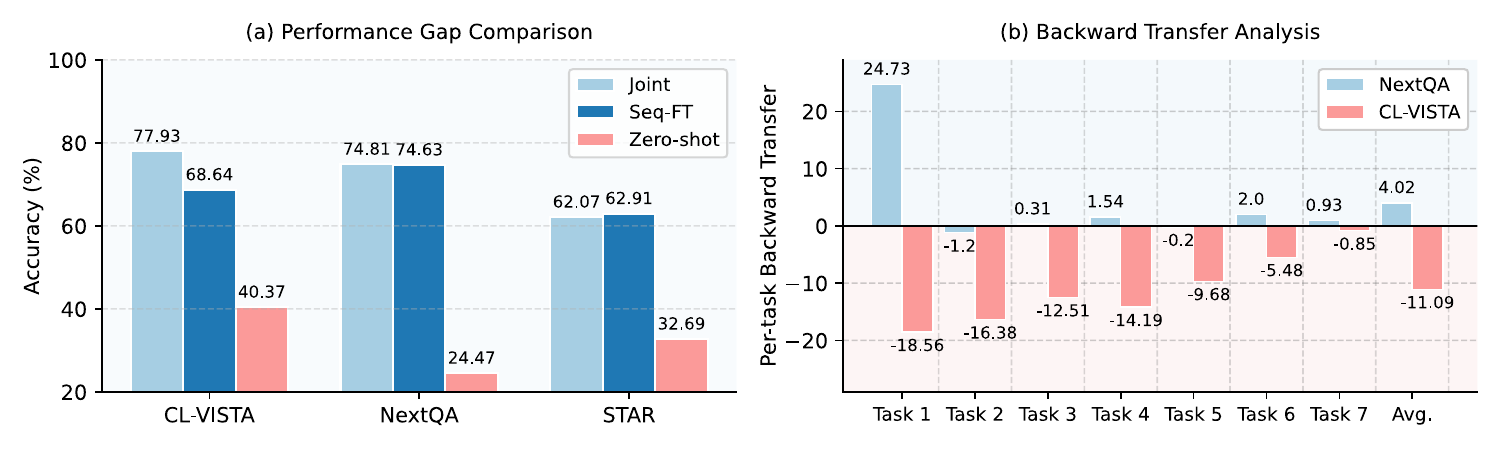}
  \vspace{-15pt}
    \caption{Comparison of CL benchmarks. (a) The significant performance gap between zero-shot and joint training indicates that these tasks require Video-LLMs to learn genuinely new knowledge. On NextQA \cite{xiao2021next} and STAR \cite{wu2024star}, sequential fine-tuning (Seq-FT) shows negligible forgetting, nearly matching the joint training upper bound. Conversely, our CL-VISTA benchmark maintains a realistic performance degradation under Seq-FT. (b) Backward Transfer (BWT) measures the impact of new learning on past tasks. Negative BWT values reveal that CL-VISTA consistently induces genuine catastrophic forgetting, whereas existing benchmarks exhibit unrealistic positive BWT.}
  \label{fig:plot_bar}
  \vspace{-15pt}
\end{figure}

To enable continual learning, Parameter-Efficient Fine-Tuning (PEFT)\cite{han2024parameter, xu2026parameter, xin2024parameter} strategies, such as LoRA\cite{hu2022lora} and prompt tuning\cite{jia2022visual}, have emerged as the predominant paradigm due to their minimal computational overhead and inherent ability to preserve pre-trained knowledge. However, despite the reported success of these methods\cite{xu2025affordance, tan2025bisecle, cai2024empowering} on existing benchmarks, we identify a significant disconnect between current continual video understanding settings and the rapid evolution of modern Video-LLMs\cite{lin2024video, cheng2024videollama}. Specifically, this mismatch stems from two primary factors: (1) \textbf{Capability Bias}, where the reliance on models (\emph{e.g.,} LLaMA-Adapter\cite{zhang2024llama}) that lack large-scale video-language pre-training deprives them of the foundational generalization required for robust video understanding; and (2) \textbf{Distribution Bias}, as prevailing benchmark protocols partition a single dataset (\emph{e.g.,} NextQA \cite{xiao2021next} or STAR \cite{wu2024star}) solely by question types, leaving the underlying video distribution static across tasks. Crucially, as illustrated in Fig. \ref{fig:plot_bar}, our empirical analysis reveals that when a truly pre-trained Video-LLM (\emph{i.e.,} Video-LLaVA \cite{lin2024video}) is employed, even sequential LoRA fine-tuning, typically regarded as a performance lower bound in CL, exhibits negligible forgetting on these benchmarks. We provide a more in-depth analysis of this phenomenon in Sec. \ref{sec:motivation}. These findings underscore that current evaluation frameworks are no longer suitable for modern Video-LLMs, as they fail to provide a sufficiently challenging environment to measure forgetting.

To bridge this gap, we introduce \textbf{CL-VISTA}, the first benchmark dedicated to \textbf{C}ontinual \textbf{L}earning of \textbf{VI}deo under\textbf{STA}nding, tailored for Video-LLMs. CL-VISTA reshapes the continual learning paradigm through two key refinements: (1) Foundation model selection, where we leverage Video-LLaVA \cite{lin2024video} and VideoLLaMA2 \cite{cheng2024videollama} to leverage their robust zero-shot capabilities derived from pre-training on large-scale video data; and (2) Downstream task construction, where we curate a diverse suite of tasks spanning perception, domain knowledge, and reasoning while rigorously preventing information leakage \cite{guo2025hide, kim2023learnability}. As illustrated in Fig. \ref{fig:plot_bar}, CL-VISTA effectively exposes the pronounced catastrophic forgetting in current Video-LLMs. To facilitate future research, we release an open-source framework that re-implements 10 mainstream multimodal CL methods, providing a unified and rigorous environment for standardized benchmarking.

Furthermore, CL-VISTA advocates for a more holistic evaluation of continual learning in Video-LLMs, moving beyond simple task-specific accuracy to establish a comprehensive framework across three critical dimensions: (1) \textbf{Performance}, where we integrate standard CL metrics with a novel general video understanding assessment to verify if models truly acquire generalized knowledge rather than simply memorizing specific tasks. Our empirical results reveal a sobering reality that most state-of-the-art methods underperform compared to simple LoRA fine-tuning in this broader context. (2) \textbf{Computational Efficiency}, which quantifies the temporal costs of the continual learning processes by measuring execution speed during both sequential training and final inference phases. (3) \textbf{Storage Overhead}, which assesses the memory footprint and parameter capacity required for practical deployment. Collectively, our findings highlight that existing methods incur prohibitive computational and storage costs to mitigate forgetting, posing significant challenges for real-world applications. The overall framework of CL-VISTA is illustrated in Fig. \ref{fig:plot_overview}. To the best of our knowledge, this is the first comprehensive effort to establish a holistic framework that integrates data, methods, and a novel evaluation system for continual learning of pre-trained Video-LLMs. We hope CL-VISTA serves as a catalyst for critical reflection and provides fresh insights into advancing CL for Video-LLMs and multimodal foundation models.

In conclusion, our main contributions are summarized as follows:

\begin{figure}[tb]
  \centering
  \includegraphics[width=\textwidth]{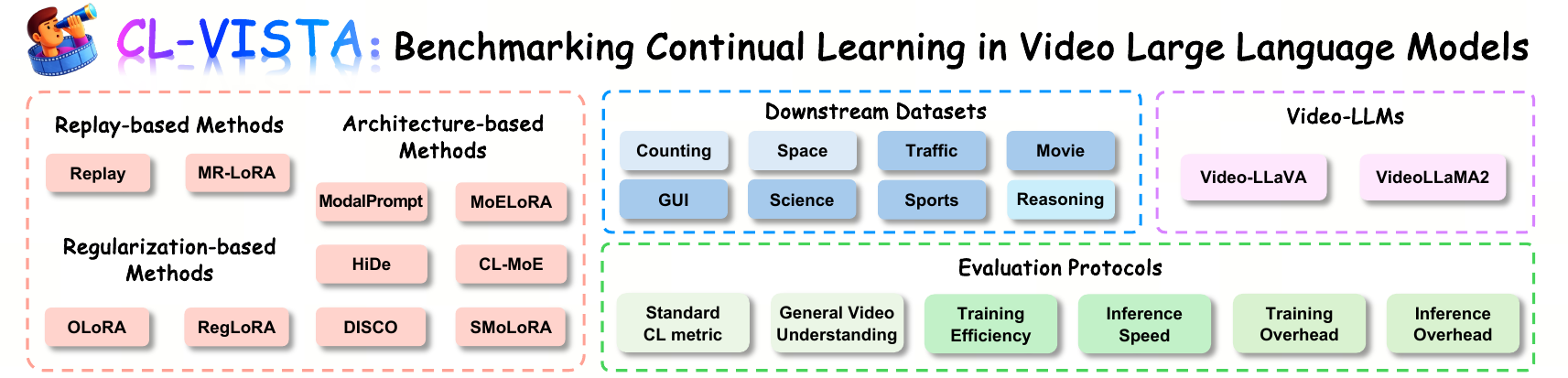}
  \vspace{-15pt}
  \caption{Overview of the CL-VISTA Benchmark: Architecture and Key Functionalities.
  }
  \label{fig:plot_overview}
  \vspace{-15pt}
\end{figure}

\begin{itemize}
    \item \textbf{CL-VISTA Benchmark.} We propose CL-VISTA, the first 
    continual learning benchmark tailored for pre-trained Video-LLMs. 
    By curating diverse tasks spanning perception, understanding, and 
    reasoning, CL-VISTA effectively reveals the catastrophic forgetting 
    that arises in current Video-LLMs under continual learning settings.
    \item \textbf{Comprehensive Evaluation Framework.} We introduce a systematic assessment protocol across three critical dimensions: performance, computational efficiency, and storage overhead. This holistic approach moves beyond standard metrics to rigorously evaluate the true capabilities and practical deployability of CL methods.
    \item \textbf{Open-source Framework and Insights.} We provide a comprehensive open-source framework that integrates mainstream multimodal CL methods. Our empirical findings reveal the generalization and efficiency bottlenecks of current state-of-the-art methods, providing critical insights for the advancement of continual learning in multimodal foundation models.
\end{itemize}
\section{Related Works}

\textbf{Continual Learning for Multimodal Large Language Models.} Existing continual learning methods for Multimodal Large Language Models (MLLMs) can be broadly categorized into replay-based methods\cite{maharana2024adapt, lee2025oasis}, regularization-based methods\cite{chen2025sefe, wang2023orthogonal, li2025multimodal}, and architecture-based methods\cite{huai2025cl, wang2025smolora, guo2025hide, zeng2025modalprompt}. Replay-based approaches mitigate catastrophic forgetting by strategically retaining representative historical samples. For instance, Adapt-$\infty$\cite{maharana2024adapt} optimizes this process via gradient-driven skill clustering and expert-led selection to curate a high-utility memory buffer. Regularization-based methods, in contrast, focus on preserving foundational competencies by imposing structural constraints. SEFE\cite{chen2025sefe} standardizes data styles using answer style diversification while employing RegLoRA to regularize key weights, ensuring the model remains robust against catastrophic interference. Architecture-based paradigms decouple learned knowledge through structural adaptation, allocating dedicated parameters to specific tasks. Specifically, HiDe-LLaVA\cite{guo2025hide} leverages layer-wise CKA similarity to implement a task-specific expansion and task-general fusion framework. Building on MoELoRA\cite{chen2024coin}, CL-MoE\cite{huai2025cl} and DISCO\cite{guo2025federated} partition task-specific knowledge into independent LoRA modules. Diverging from traditional routing, they employ multimodal prototype routers to selectively activate experts, effectively safeguarding prior knowledge. Furthermore, MR-LoRA\cite{zhao2025mllm} integrates a native MLLM-based router trained with sparse replay data, achieving an effective balance between stability and plasticity. However, the aforementioned methods are predominantly developed for image-text MLLMs, and whether they remain effective for Video-LLMs, where temporal understanding and reasoning introduce fundamentally different challenges, remains largely unexplored.

\noindent
\textbf{Continual Learning for Video Understanding.} To handle diverse video understanding tasks in a continual learning manner, recent studies\cite{cheng2025dam, tan2025bisecle, cai2024empowering, xu2025affordance, tang2024vilco, park2021class, chawla2024continual} have explored various techniques to mitigate catastrophic forgetting in video-based models. From a methodological perspective, various parameter-efficient tuning strategies have been explored to preserve knowledge. Specifically, DAM \cite{cheng2025dam} aggregates sequential expertise through a dynamic adapter merging scheme. ColPro \cite{cai2024empowering} utilizes collaborative prompting to capture the intricate temporal dynamics of VideoQA. Parallel to these, Bisecle \cite{tan2025bisecle} introduces a bio-inspired approach that mimics hippocampal binding and separation, using contrastive prompts to mitigate inter-task interference. Despite their progress, these frameworks lack native Video-LLM integration, making them incompatible with the evolving foundational capabilities of large-scale video modeling.
In terms of evaluation benchmarks, VilCo-Bench\cite{tang2024vilco} provides a systematic evaluation framework for video-text tasks, extending beyond traditional class-incremental settings with long-duration videos. Cai et al.\cite{cai2024empowering} categorize tasks based on VideoQA types, establishing a taxonomy that has been widely adopted in subsequent research\cite{tan2025bisecle, xu2025affordance}. However, our empirical findings reveal that such task-specific partitioning is ill-suited for native Video-LLMs, limiting their potential in advanced video understanding. Consequently, there is a pressing need for a benchmark and framework built upon native Video-LLMs to fully leverage their foundational capabilities in continual video understanding.
\section{CL-VISTA: A Continual Learning Benchmark for Video-LLMs}

\subsection{Problem Definition}

CL-VISTA is tailored for continual learning in Video Question Answering. Formally, let $\mathcal{S} = \{ \mathcal{D}_t \}_{t=1}^T$ be a sequence of $T$ tasks. Each task $\mathcal{D}_t$ contains $n_t$ samples, where each sample $d_i^t \in \mathcal{D}_t$ is represented as a triplet $(V_i^t, Q_i^t, A_i^t)$. Here, $V_i^t$ denotes the video clip, $Q_i^t$ the question, and $A_i^t$ the target answer. In a conventional Supervised Fine-Tuning~(SFT) paradigm, the Video-LLM is trained to predict the next token in an auto-regressive manner. For the $t$-th task $\mathcal{D}_t$, the training objective is to optimize the model parameters $\theta_t$ by minimizing the negative log likelihood loss:
\begin{equation}
    \mathcal{L}_{\text{SFT}}(\theta_t) = - \sum_{i=1}^{n_t} \sum_{l=1}^{L_i} \log P_{\theta_t} (a_{i,l}^t | V_i^t, Q_i^t, a_{i,<l}^t),
\end{equation}
where $L_i$ is the sequence length of the answer $A_i^t$, and $a_{i,l}^t$ denotes the $l$-th token of the target answer. The term $a_{i,<l}^t$ represents the tokens preceding the $l$-th position. In the context of continual learning, the model aims to optimize $\mathcal{L}_{\text{SFT}}$ on the current task $\mathcal{D}_t$ while preventing performance degradation on previously learned tasks. Formally, the goal is to maintain the model's proficiency on the historical task stream $\mathcal{S}_{<t} = \{\mathcal{D}_1, \dots, \mathcal{D}_{t-1}\}$ without having direct access to full datasets, thereby mitigating the effects of catastrophic forgetting.

\begin{figure}[h]
  \centering
  \includegraphics[width=\textwidth]{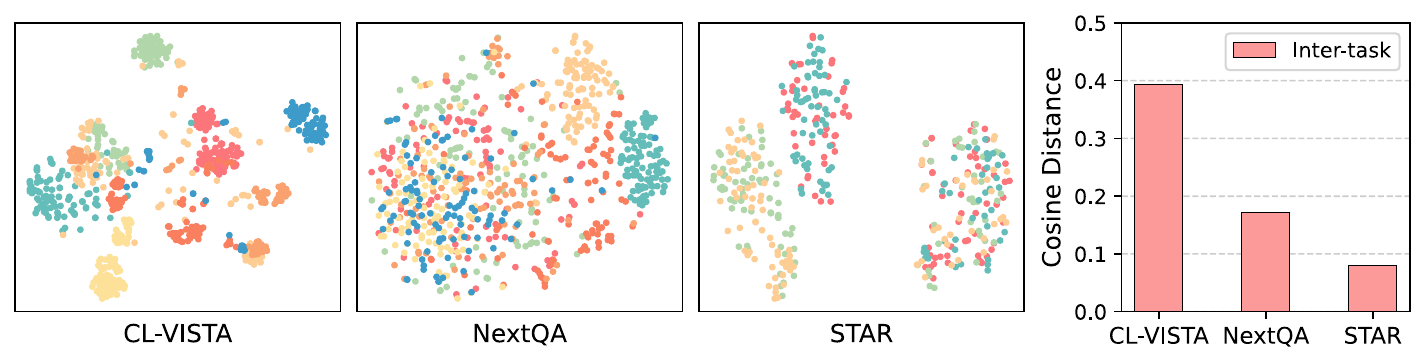}
  \caption{Embedding discriminability analysis. Compared to previous benchmarks where task embeddings are highly entangled, our proposed setting yields clearer task separation. Quantitative results (right) further confirm that our benchmark achieves significantly larger inter-task distances, facilitating better task-incremental evaluation.
  }
  \label{fig:plot_tsne}
\end{figure}

\subsection{Motivation}
\label{sec:motivation}

The unexpected resilience against catastrophic forgetting shown in Fig. \ref{fig:plot_bar} implies that current benchmarks are insufficient to challenge the continual learning capabilities of Video-LLMs. We contend that this phenomenon stems from inherent limitations in benchmark design, manifested as Distribution Bias and Capability Bias (Sec.~\ref{sec:intro}). To validate this hypothesis, we link these two concepts to our empirical analysis through the following two key perspectives.

\noindent
\textbf{Embedding discriminability analysis.} To evaluate semantic separation across tasks, we visualize the joint video-question representation space via t-SNE \cite{maaten2008visualizing} on embeddings extracted by Qwen3-VL-Embedding \cite{li2026qwen3}. As shown in Fig. \ref{fig:plot_tsne}, embeddings from existing benchmarks collapse into a monolithic cluster without discernible semantic boundaries, indicating severe inter-task entanglement. For rigorous quantification, we measure the inter-task cosine distance based on task centroids, where each centroid is computed as the L2-normalized mean of all sample embeddings within a task. The inter-task distance is then defined as the mean pairwise cosine distance across all task centroids, with higher values reflecting reduced redundancy and greater semantic diversity. Quantitative results demonstrate that CL-VISTA achieves substantially larger inter-task separation compared to existing benchmarks, confirming that our dataset exposes models to genuinely distinct task distributions and effectively addresses the \emph{distribution bias} prevalent in prior work.

\begin{figure}[h]
  \centering
  \includegraphics[width=\textwidth]{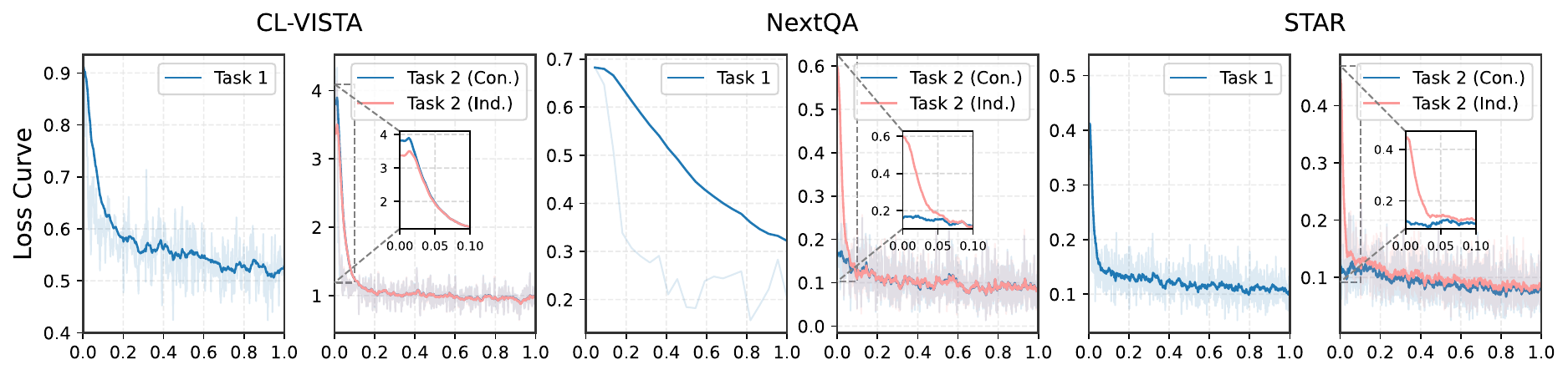}
  \caption{Learning trajectory analysis. Compared to individual training (\textcolor{mypink}{pink}), CL-VISTA’s continual training (\textcolor{myblue}{blue}) exhibits distinct loss spikes at task boundaries. Existing benchmarks show smooth, homogeneous transitions, failing to challenge Video-LLMs with the clear distribution shifts necessary for evaluating continual learning.
  }
  \label{fig:plot_loss}
\end{figure}

\noindent
\textbf{Learning trajectory analysis.} The optimization dynamics further corroborate these findings. In Fig. \ref{fig:plot_loss}, we plot the training loss of Video-LLaVA (with LoRA) for the second task under two settings: individual training and continual training. For existing benchmarks, the sequential loss begins at a level nearly identical to the convergence point of the previous task, which is significantly lower than its individual counterpart. Such a warm-start phenomenon implies that the model perceives the new task as a mere continuation of previous ones, highlighting a high degree of task redundancy in current benchmarks. In contrast, CL-VISTA exhibits a starting loss that is even higher than the standalone baseline, indicating a substantial distribution shift. This confirms that CL-VISTA effectively mitigates \emph{capability bias} by forcing the model to undergo genuine knowledge acquisition, thereby providing a more rigorous and authentic testbed for catastrophic forgetting.

\begin{table}[h]
  \vspace{5pt}
  \caption{Overview of downstream tasks in the CL-VISTA benchmark. Tasks are categorized into Perception (Per.), Understanding (Und.), and Reasoning (Rea.). ``Recon.'' indicates whether the task is curated from original video and annotation sources (\xmark) or involves data reconstruction (\cmark).}
  \label{tab:vista_tasks}
  \centering
  \resizebox{\textwidth}{!}{ 
  \setlength{\aboverulesep}{0pt}
  \setlength{\belowrulesep}{0pt}
  \renewcommand{\arraystretch}{1.2} 
  \setlength{\tabcolsep}{5pt}
  \begin{tabular}{clllccc}
    \toprule
    \textbf{Dimension} & \textbf{Task Name} & \textbf{Video source} & \textbf{Annotation source} & \textbf{Train size} & \textbf{Test size} & \textbf{Recon.} \\
    \midrule
    \multirow{2}{*}{Per.} 
    & Counting & Molmo2-Pointing\cite{clark2026molmo2} & Molmo2-Pointing\cite{clark2026molmo2} & 20,000 & 2,000 & \xmark \\
    & Spatial  & ScanNet\cite{dai2017scannet} & Spatial-MLLM\cite{wu2025spatial} & 30,000 & 2,000 & \xmark \\
    \hdashline
    \multirow{5}{*}{Und.}
    & Traffic   & \makecell[l]{TUMTraffic\cite{zhou2025tumtraffic} \\ RoadSocial\cite{parikh2025roadsocial} \\ LingoQA\cite{marcu2024lingoqa}} & \makecell[l]{TUMTraffic\cite{zhou2025tumtraffic} \\ RoadSocial\cite{parikh2025roadsocial} \\ LingoQA\cite{marcu2024lingoqa}} & 30,000 & 2,000 & \xmark \\
    & Movie       & CinePile\cite{rawal2024cinepile} & CinePile\cite{rawal2024cinepile} & 30,000 & 2,000 & \xmark \\
    & GUI       & GUI-World\cite{chen2024gui} & GUI-World\cite{chen2024gui} & 30,000 & 2,000 & \xmark \\
    & Science        & FineVideo\cite{FineVideo} & FineVideo\cite{FineVideo} & 30,000 & 2,000 & \cmark \\
    & Sports    & \makecell[l]{Sports-QA\cite{li2024sports} \\ SoccerChat\cite{Gautam2025May}} & \makecell[l]{Sports-QA\cite{li2024sports} \\ SoccerChat\cite{Gautam2025May}} & 30,000 & 2,000 & \cmark \\
    \hdashline
    Rea. & Reasoning & STAR\cite{wu2024star} & STAR\cite{wu2024star} & 25,000 & 2,000 & \xmark \\
  \bottomrule
  \end{tabular}
  }
\end{table}

\subsection{Benchmark Setup}

\textbf{Data Integration.} CL-VISTA is designed as a comprehensive continual video understanding benchmark spanning three core dimensions: perception, understanding, and reasoning. The \textit{perception} dimension comprises counting and spatial perception tasks; the \textit{understanding} dimension encompasses five vertical domains, including traffic, movies, GUI, science, and sports; and the \textit{reasoning} dimension focuses on temporal reasoning capabilities. We maximize the utility of open-source datasets through both direct integration and secondary reconstruction, as summarized in Table~\ref{tab:vista_tasks}.

\begin{figure}[h]
  \centering
  \includegraphics[width=\textwidth]{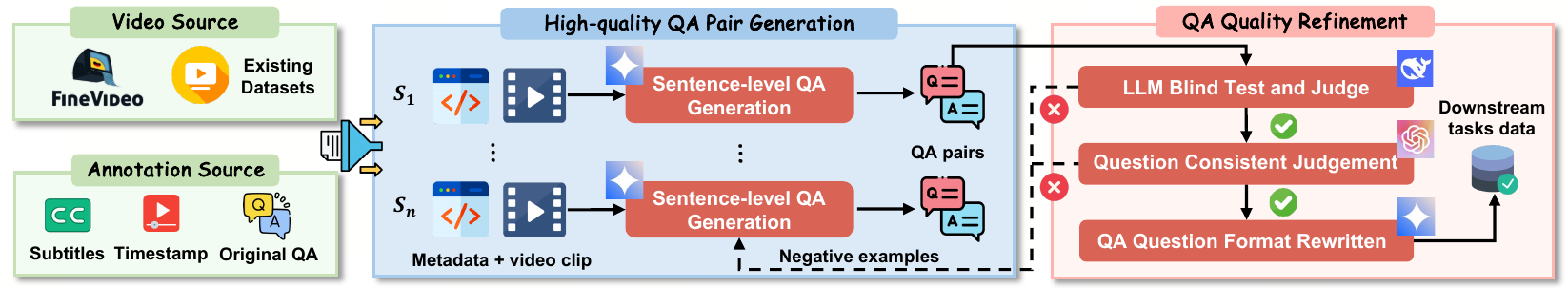}
  \caption{
    \textbf{Overview of the data reconstruction pipeline.} 
    The pipeline begins with the generation of sentence-level 
    question-answer pairs derived from segmented video clips 
    and their corresponding annotations. These candidates are 
    subsequently refined through a multi-discriminator filtering 
    mechanism to yield a final collection of high-quality and 
    diverse QA pairs.}
  \label{fig:data-pipeline}
\end{figure}

To address the scarcity of off-the-shelf training data for specialized domains like science and sports, we develop a high-quality Video-QA generation pipeline (Fig. \ref{fig:data-pipeline}). We first perform temporal segmentation on long videos based on metadata and timestamps to extract semantically coherent clips. These segments, along with their subtitles, are processed by a generator to produce standardized, single-sentence QA pairs. To ensure data integrity, we employ a dual-discriminator system for rigorous validation: (1) a blind test to identify modality-biased samples that are answerable without video context, and (2) a consistency check to verify the alignment between questions and answers. Samples flagged as negatives are fed back into the generator for refinement. Finally, validated QA pairs are rewritten to introduce stylistic diversity, forming the final downstream task datasets. Further details are provided in Appendix \ref{appendix_A}.

\noindent
\textbf{Model Configuration.} We employ Video-LLaVA \cite{lin2024video} and VideoLLaMA2 \cite{cheng2024videollama} as the foundational backbones for CL-VISTA. Unlike early architectures (e.g., LLaMA-Adapter \cite{zhang2024llama}) that lack large-scale video alignment, these contemporary Video-LLMs benefit from extensive video-centric pre-training. Consequently, they possess robust foundational video understanding capabilities, providing a realistic testbed to investigate catastrophic forgetting and ensuring our findings readily generalize to state-of-the-art multimodal models.

\noindent
\textbf{Supported Methods.} CL-VISTA benchmarks 10 representative continual learning methods across two distinct Video-LLM architectures, including Replay \cite{rebuffi2017icarl}, OLoRA \cite{wang2023orthogonal}, MoELoRA \cite{chen2024coin}, ModalPrompt \cite{zeng2025modalprompt}, RegLoRA \cite{chen2025sefe}, CL-MoE \cite{huai2025cl}, HiDe \cite{guo2025hide}, DISCO \cite{guo2025federated}, SMoLoRA \cite{wang2025smolora}, and MR-LoRA \cite{zhao2025mllm}. While these methods have shown strong performance on image-text MLLMs, their effectiveness on Video-LLMs, where temporal reasoning introduces fundamentally different demands, remains an open question. Detailed descriptions and hyperparameter configurations are provided in Appendix \ref{appendix_B}.

\noindent
\textbf{Evaluation Protocol.} To systematically assess the effectiveness and practicality of continual learning methods, we establish a comprehensive evaluation framework comprising 6 distinct protocols categorized into three primary dimensions: performance, computational efficiency, and storage overhead. For the performance dimension, we evaluate different methods through standard CL metrics and general video understanding. Following prior work \cite{chen2025sefe}, the standard CL evaluation reports Mean Finetune Accuracy (MFT), Mean Final Accuracy (MFN), Mean Average Accuracy (MAA), and Backward Transfer (BWT), with detailed mathematical definitions provided in Appendix \ref{appendix_C}. Notably, we employ Qwen3-30A3B-Instruct \cite{yang2025qwen3} as an automated judge to mitigate spurious forgetting \cite{chen2025sefe}, a phenomenon where performance degrades due to shifts in instruction-following behavior rather than genuine knowledge loss. To evaluate general video understanding, we further assess performance on a suite of benchmarks, including MMVU \cite{zhao2025mmvu}, MVBench \cite{li2024mvbench}, NExTQA \cite{xiao2021next}, LongVideoBench \cite{wu2024longvideobench}, and MMBench-Video \cite{fang2024mmbench}. This investigates whether existing CL methods truly facilitate knowledge integration across the entire task sequence or merely lead to task-specific overfitting, ensuring the model maintains its fundamental video understanding capabilities in a general-purpose context. Beyond performance, computational efficiency is evaluated under identical environments, measuring training efficiency as the average execution time across sequential tasks and inference efficiency as the time required to evaluate all tasks using the final model. Finally, storage overhead assesses the memory footprint of these inherently LoRA-based methods. Specifically, training overhead is quantified by the parameter size required to train the final task, while inference overhead measures the total weight capacity needed to evaluate all encountered tasks. In summary, this framework provides a holistic assessment of whether a CL algorithm can be practically deployed in resource-constrained real-world scenarios, rather than solely optimizing for performance on a fixed set of downstream tasks.

\section{Experiments}

\subsection{Implementation details}

We implement the baseline methods by leveraging their official source code and the open-source continual instruction tuning codebase\cite{guo2025mcitlib}. To ensure a fair comparison, all method-specific hyperparameters are kept consistent with the default settings reported in their original papers. All experiments are conducted using Video-LLaVA-7B and VideoLLaMA2-7B as the backbone models, with 8 frames uniformly sampled from each video by default. During the training phase, each task is optimized for one epoch with a global batch size of 64 using DeepSpeed Zero-2\cite{rasley2020deepspeed} on 8$\times$A100 GPUs. The learning rates for PEFT and projection layers are set to $1 \times 10^{-4}$ and $1 \times 10^{-5}$, respectively, with a warmup ratio of 0.03. For replay-based methods, we fix the number of replay samples at 100 per task. The task sequence for all experiments is defined as: \textit{Counting} $\rightarrow$ \textit{Space} $\rightarrow$ \textit{Traffic} $\rightarrow$ \textit{Movie} $\rightarrow$ \textit{GUI} $\rightarrow$ \textit{Science} $\rightarrow$ \textit{Sports} $\rightarrow$ \textit{Reasoning}. Detailed hyperparameters are provided in Appendix \ref{appendix_B}.

\begin{table*}[h]
\vspace{5pt}
    \centering
    \caption{Performance comparison of different methods on Video-LLaVA using standard continual learning metrics. The best results are highlighted in \textbf{bold}, and the second-best results are \underline{underlined}.}
    \label{tab:standard_cl}
    \renewcommand{\arraystretch}{1.1} 

    \resizebox{\linewidth}{!}{
    \begin{tabular}{@{} l *{8}{c} *{4}{c} @{}}
        \toprule
        Method & 
        \vhead{Count.} & \vhead{Space} & \vhead{Traffic} & \vhead{Movie} & 
        \vhead{GUI} & \vhead{Science} & \vhead{Sports} & \vhead{Reason.} & 
        \phead{MFT ($\uparrow$)} & \phead{MFN ($\uparrow$)} & \phead{MAA ($\uparrow$)} & \phead{BWT ($\uparrow$)} \\ 
        \midrule
        Zero-shot  & 32.24 & 37.24 & 41.42 & 22.78 & 60.17 & 56.91 & 34.40 & 37.52 & 40.34 & -- & -- & -- \\
        Individual & 59.69 & 72.81 & 67.04 & 89.63 & 79.07 & 83.43 & 89.75 & 89.43 & 78.86 & -- & -- & -- \\
        Joint & 60.20 & 67.07 & 70.13 & 86.99 & 78.11 & 83.97 & 90.98 & 85.98 & 77.93 & -- & -- & -- \\
        LoRA-FT    & 41.13 & 54.81 & 55.84 & 74.61 & 68.80 & 78.20 & 88.90 & 86.86 & 78.35 & 68.64 & 63.55 & -11.09 \\
        \midrule
        \rowcolor{myblue!10} \multicolumn{13}{c}{\textbf{Replay-based Methods}} \\
        Replay     & 53.39 & 55.13 & 59.27 & 74.78 & 70.91 & 79.67 & 86.7 & 87.76 & 78.12 & 70.95 & 65.77 & -8.19 \\
        MR-LoRA      & \textbf{59.72} & \underline{72.50} & \textbf{67.99} & \underline{89.85} & \textbf{79.46} & \textbf{82.89} & \textbf{89.46} & \underline{89.75} & \textbf{78.94} & \textbf{78.95} & \textbf{71.34} & \textbf{0.02} \\
        \midrule
        \rowcolor{myblue!10} \multicolumn{13}{c}{\textbf{Regularization-based Methods}} \\
        OLoRA     & 41.05 & 52.53 & 41.00 & 58.82 & 57.94 & 62.37 & 67.46 & 46.97 & 64.26 & 53.52 & 61.39 & -17.06 \\
        RegLoRA   & 45.20 & 43.47 & 46.89 & 37.23 & 59.17 & 60.36 & 52.92 & 86.51 & 76.20 & 53.97 & 52.76 & -25.41 \\
        \midrule
        \rowcolor{myblue!10} \multicolumn{13}{c}{\textbf{Architecture-based Methods}} \\
        MoELoRA    & 40.44 & 52.88 & 54.27 & 72.46 & 67.27 & 79.29 & 87.47 & 84.23 & 76.36 & 67.29 & 62.88 & -10.37 \\
        ModalPrompt& 31.81 & 26.32 & 41.38 & 38.14 & 41.55 & 56.20 & 29.20 & 36.76 & 39.77 & 37.67 & 35.14 & -2.40 \\
        CL-MoE     & 42.79 & 55.77 & 54.15 & 74.28 & 67.65 & 78.06 & 87.79 & 86.58 & 77.45 & 68.38 & 63.42 & -10.36 \\
        HiDe       & 41.35 & 47.32 & 39.01 & 64.46 & 60.29 & 64.36 & 67.46 & 51.26 & 66.61 & 54.44 & 60.15 & -13.91 \\
        SMoLoRA    & 51.66 & 59.45 & 57.70 & 80.64 & 71.17 & \underline{81.87} & 87.31 & 86.92 & \underline{78.18} & 72.09 & 67.27 & -6.95 \\
        DISCO      & \underline{59.69} & \textbf{72.58} & \underline{61.18} & \textbf{90.45} & \underline{78.27} & 78.61 & \textbf{88.16} & \textbf{90.04} & 78.12 & \underline{77.37} & \underline{70.31} & \underline{-0.85} \\
        \bottomrule
    \end{tabular}}
\end{table*}

\subsection{Performance on Standard CL Metrics}

Table \ref{tab:standard_cl} reports the performance of various continual learning methods on the Video-LLaVA model. Our empirical results reveal a significant polarization in the effectiveness of existing multimodal CL methods when applied to Video-LLMs:

\textbf{(1) Severe forgetting in existing methods.} Many established CL approaches fail to effectively mitigate forgetting, with some even underperforming the LoRA-FT baseline. For instance, RegLoRA and HiDe yield BWT scores of -25.41\% and -13.91\%, worse by 14.32 and 2.82 points than LoRA-FT (-11.09\%), respectively. Meanwhile, architecture-based methods such as MoELoRA and CL-MoE achieve BWT scores of -10.37\% and -10.36\%, barely on par with the LoRA-FT baseline, indicating that simply scaling model capacity offers negligible benefit for forgetting mitigation. We attribute these failures primarily to the high heterogeneity of the CL-VISTA benchmark, which spans a wide range of capabilities from basic perception to temporal reasoning. In such a diverse task sequence, static regularization constraints, generic knowledge fusion mechanisms, and undifferentiated mixture-of-experts architectures struggle to adapt to the drastic shifts in task-specific requirements.

\textbf{(2) Superiority of task-specific adaptation methods.} In contrast, methods that assign dedicated parameters or data to individual tasks demonstrate a stronger capability to alleviate forgetting. Among architecture-based methods, DISCO leverages parameter isolation by allocating task-specific modules and invoking them via textual-feature matching during inference, achieving a BWT of -0.85\%. More notably, MR-LoRA, a rehearsal-based approach, achieves the best overall performance across all aggregated metrics (MFN: 78.95, MAA: 71.34, BWT: 0.02), outperforming all other methods, including DISCO. Rather than relying on external similarity metrics, MR-LoRA utilizes a small amount of stored data to train the model itself as a native router, enabling autonomous selection of appropriate expert parameters for each input. By effectively decoupling the learning process into task-specific components, both methods preserve specialized knowledge across diverse video contexts.

Experiments on VideoLLaMA2 (Appendix \ref{appendix_D}) further validate these findings across different architectures. Nevertheless, the strong performance of task-specific adaptation methods on standard metrics raises a critical question: do they genuinely generalize to unseen tasks, or merely fit the observed task distributions? We investigate this in the next section.

\begin{table*}[h]
\vspace{5pt}
    \centering
    \caption{General video understanding performance of various CL methods on Video-LLaVA. We highlight the relative change compared to the LoRA-FT baseline (\textcolor{upcolor}{green} for improvement, \textcolor{downcolor}{red} for decline).}
    \label{tab:general_video}
    \renewcommand{\arraystretch}{1.1}
    \setlength{\tabcolsep}{5pt}

    \resizebox{0.98\linewidth}{!}{
    \begin{tabular}{@{} l ccccc c @{}}
        \toprule
        \textbf{Method} & \textbf{MMVU} & \textbf{MVBench} & 
        \makecell[c]{\textbf{LongVideo}\\\textbf{Bench}} & 
        \makecell[c]{\textbf{MMBench-}\\\textbf{Video}} & \textbf{NExTQA} & \textbf{Average} \\
        \midrule
        Zero-shot  & 28.60 & 41.54 & 37.70 & 31.78 & 23.94 & 32.71 \\
        Joint      & 36.60 & 45.68 & 42.71 & 34.66 & 53.50 & 42.63 \\
        LoRA-FT    & 34.80 & 46.86 & 45.03 & 34.03 & 58.77 & 43.90 \\
        \midrule
        \rowcolor{myblue!10} \multicolumn{7}{c}{\textbf{Replay-based Methods}} \\
        Replay     & 37.10 \up{2.30} & 48.03 \up{1.17} & 44.65 \down{0.38} & 33.94 \down{0.09} & 59.61 \up{0.84} & 44.67 \up{0.77} \\
        MR-LoRA    & 30.90 \down{3.90} & 46.78 \down{0.08} & 40.80 \down{4.23} & 33.51 \down{0.52} & 42.79 \down{15.98} & 38.96 \down{4.94} \\
        \midrule
        \rowcolor{myblue!10} \multicolumn{7}{c}{\textbf{Regularization-based Methods}} \\
        OLoRA      & 18.00 \down{16.80} & 33.87 \down{12.99} & 32.54 \down{12.49} & 19.30 \down{14.73} & 20.43 \down{38.34} & 24.83 \down{19.07} \\
        RegLoRA    & 28.10 \down{6.70} & 47.89 \up{1.03} & 39.12 \down{5.91} & 34.03 \up{0.00} & 44.42 \down{14.35} & 38.71 \down{5.19} \\
        \midrule
        \rowcolor{myblue!10} \multicolumn{7}{c}{\textbf{Architecture-based Methods}} \\
        MoELoRA    & 33.30 \down{1.50} & 46.63 \down{0.23} & 43.90 \down{1.13} & 33.52 \down{0.51} & 53.62 \down{5.15} & 42.19 \down{1.71} \\
        ModalPrompt& 22.70 \down{12.10} & 37.18 \down{9.68} & 36.57 \down{8.46} & 30.32 \down{3.71} & 24.10 \down{34.67} & 30.17 \down{13.73} \\
        CL-MoE     & 33.20 \down{1.60} & 47.07 \up{0.21} & 45.18 \up{0.15} & 33.49 \down{0.54} & 61.49 \up{2.72} & 44.09 \up{0.19} \\
        HiDe       & 22.30 \down{12.50} & 36.08 \down{10.78} & 34.26 \down{10.77} & 20.28 \down{13.75} & 24.26 \down{34.51} & 27.44 \down{16.46} \\
        SMoLoRA    & 36.20 \up{1.40} & 47.47 \up{0.61} & 44.88 \down{0.15} & 33.74 \down{0.29} & 57.49 \down{1.28} & 43.96 \up{0.06} \\
        DISCO      & 31.90 \down{2.90} & 43.74 \down{3.12} & 42.48 \down{2.55} & 33.82 \down{0.21} & 37.65 \down{21.12} & 37.92 \down{5.98} \\
        \bottomrule
    \end{tabular}}
\end{table*}

\subsection{Performance on General Video Understanding}

To investigate whether success on CL-VISTA translates into broader generalization, we conduct a comprehensive evaluation on several general video understanding benchmarks. Specifically, we use the final checkpoints of each method to assess their performance on 5 mainstream benchmarks: NExTQA\cite{xiao2021next}, MMVU\cite{zhao2025mmvu}, MVBench\cite{li2024mvbench}, LongVideoBench\cite{wu2024longvideobench}, and MMBench-Video\cite{fang2024mmbench}. Notably, these benchmarks have no data overlap with our training set. The results on Video-LLaVA are presented in Table \ref{tab:general_video}.

First, joint training on CL-VISTA significantly improves general video understanding performance compared to the Zero-shot baseline. Notably, LoRA-FT (49.99\%) achieves a higher average score than Joint training (49.04\%). This outcome may stem from the inherent task interference in joint training, where randomly mixing 8 diverse domains can lead to conflicting gradient updates. In contrast, the sequential training in LoRA-FT, which transitions from foundational perception to intermediate understanding and finally to complex reasoning, aligns with the principles of curriculum learning by presenting tasks in an increasing order of difficulty\cite{wang2021survey}. This progressive increase in task complexity facilitates the development of more robust representations, suggesting that while LoRA-FT exhibits forgetting in standard CL metrics, it excels in fostering generalized capabilities.

We further examine whether specialized CL baseline methods can surpass the generalized performance of LoRA-FT. As shown in Table \ref{tab:general_video}, the results indicate a significant gap between traditional CL success and universal video understanding. Specifically, methods like RegLoRA and HiDe exhibit consistently poor results, failing to adapt to the complex temporal dynamics of video tasks; HiDe, in particular, suffers a catastrophic 16.46\% average decline compared to LoRA-FT, with its NExTQA performance plummeting by 34.51\%. Crucially, a performance-forgetting paradox emerges in state-of-the-art methods such as DISCO and MR-LoRA. Despite their efficacy in mitigating catastrophic forgetting on seen tasks, these methods exhibit substantial average performance decrements of 5.98\% and 4.94\% on CL-VISTA. This discrepancy suggests that while their sophisticated routing mechanisms successfully preserve task-specific knowledge within observed distributions, they simultaneously induce distributional over-specialization. Such architectural rigidity constrains the model’s plasticity, thereby impeding the effective transfer and generalization of acquired representations to broader, unseen video domains. This phenomenon is particularly evident in NExTQA, where DISCO and MR-LoRA decline by 21.12\% and 15.98\%, respectively, indicating that current CL paradigms may prioritize memory retention at the expense of universal knowledge transferability.

In summary, results across architectures, with VideoLLaMA2 results in Appendix D, show a trade-off between stability and plasticity. Current methods either sacrifice generalization to prevent forgetting or fail to provide stability altogether. Therefore, while a practical CL system must achieve both, this remains a major challenge for existing frameworks.

\begin{figure}[h]
  \centering
  \includegraphics[width=\textwidth]{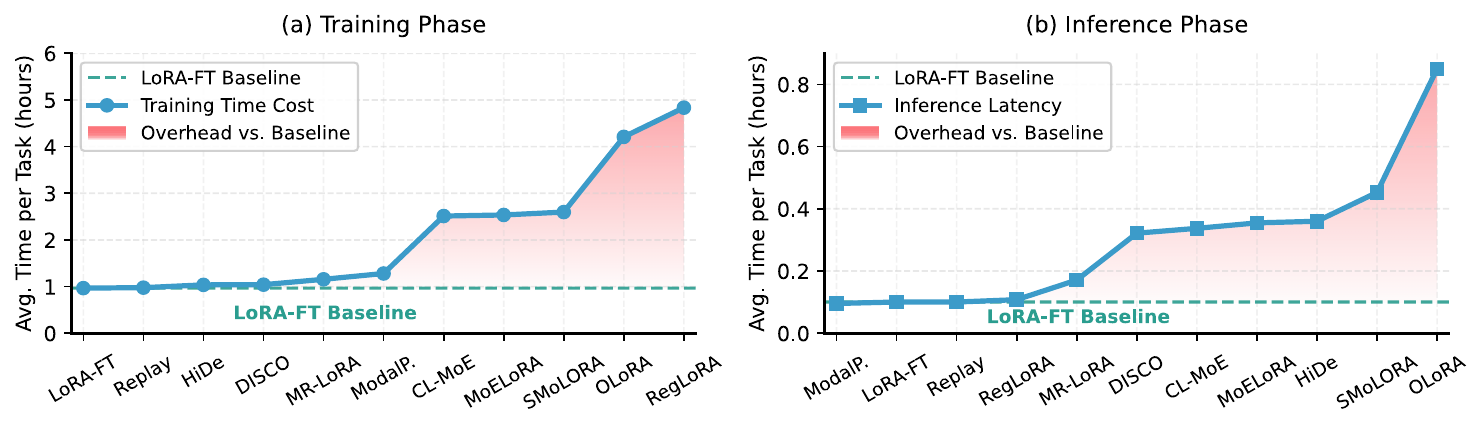}
  \caption{
  Comparative analysis of computational efficiency among various LoRA-based continual learning methods on Video-LLaVA. (a) Average training time cost per task. (b) Average inference latency per task. The dashed line represents the standard LoRA-FT lower bound. The red shaded regions explicitly highlight the additional computational delay introduced by each specific CL method relative to the baseline. Methods are sorted in ascending order of their time costs.
  }
  \label{fig:plot_speed}
\end{figure}

\subsection{Training and Inference Efficiency}

In addition to predictive accuracy, the computational efficiency of CL methods is a critical factor for practical deployment. To ensure an equitable comparison, all runtime evaluations based on Video-LLaVA are conducted under strictly identical environment configurations. Training efficiency is quantified by the average execution time across sequential tasks, with timing metrics extracted from the \texttt{trainer\_state.json} logs of the model checkpoints. Inference efficiency is determined by the average time required to evaluate all tasks using the final trained model. All temporal measurements are normalized to hours and presented in Fig \ref{fig:plot_speed}.

Regarding training efficiency, as illustrated in Fig. \ref{fig:plot_speed}(a), regularization-based approaches such as RegLoRA and OLoRA exhibit the most substantial computational overhead. These methods require approximately 4 additional hours per task compared to the LoRA-FT baseline. Such extensive delays become prohibitive in continual learning scenarios involving long sequences of tasks, particularly given that these approaches do not yield commensurate improvements in standard CL metrics or general video benchmarks. We attribute this severe training latency to suboptimal infrastructure within their current implementations. Specifically, there is a lack of integration between their custom regularization computations and standard acceleration frameworks, such as DeepSpeed, which are typically employed for video-LLMs. Furthermore, a similar trend of computational inefficiency is observed in architecture-expansion methods. Approaches including SMoLoRA and MoELoRA, which necessitate the dynamic adjustment and allocation of Mixture-of-Experts routing during training, introduce considerable computational burdens that severely hinder their practical applicability.

During the inference phase, as shown in Fig. \ref{fig:plot_speed}(b), latency issues remain prominent and are particularly severe among architecture-expansion methods. While approaches like DISCO achieve strong empirical results on standard continual learning metrics, this performance comes at the cost of a nearly threefold increase in inference time relative to the baseline. Other methods, including SMoLoRA and OLoRA, demonstrate even higher inference delays. This substantial increase in inference latency highlights a critical trade-off between predictive accuracy and computational efficiency. Consequently, the heavy computational demands of these high-performing methods pose significant challenges for the real-time deployment of continual learning models in practical applications.

\begin{figure}[h]
  \centering
  \includegraphics[width=\textwidth]{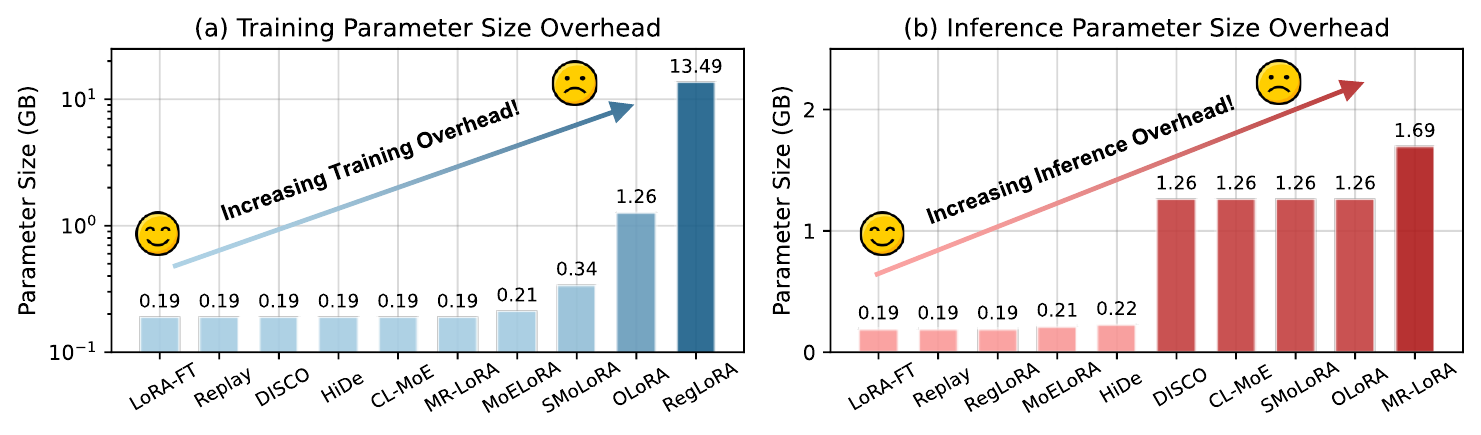}
  \caption{
  Comparison of parameter size overhead (GB) among ten 
continual learning methods during (a) training and (b) inference 
phases. Methods are sorted in ascending order of overhead. 
Darker colors indicate larger parameter size consumption. 
Note that (a) uses a logarithmic scale due to the significant 
overhead introduced by RegLoRA.
  }
  \label{fig:plot_overhead}
\end{figure}

\subsection{Storage Overhead Analysis}

Beyond computational efficiency, the storage overhead incurred during both the training and inference phases is a critical consideration for the practical deployment of CL methods, particularly in resource-constrained edge environments. Accordingly, we evaluate the memory footprint of various approaches on Video-LLaVA across both phases, with the comparative results illustrated in Fig. \ref{fig:plot_overhead}. To quantify the training overhead, we calculate the total size of the parameters required during the training of the final task. For the inference overhead, we measure the total capacity of the model weights necessary to evaluate the test sets of all encountered tasks following the completion of the entire training sequence. Notably, all compared methods are inherently LoRA-based methods.

During the training phase, Fig. \ref{fig:plot_overhead}(a) reveals that the majority of methods maintain a relatively low memory footprint, with the notable exception of regularization-based approaches. Specifically, to compute the regularization loss during the training of the final task, RegLoRA necessitates loading the key parameters from all previously encountered tasks into memory, thereby imposing a substantial storage burden. Similarly, OLoRA requires iterating over the parameter weights of all prior tasks to calculate its orthogonal loss. Consequently, these exorbitant memory requirements indirectly contribute to the severe training delays associated with such methods.

During the inference phase, most parameter-expansion methods exhibit a pronounced increase in memory consumption. This is expected, as their fundamental design paradigm relies on trading spatial efficiency for improved performance. These approaches typically allocate independent LoRA modules for each task during training and dynamically load the corresponding weights during inference based on specific routing rules. Consequently, to effectively mitigate catastrophic forgetting, they are compelled to retain the parameter weights of all historical tasks. Among these, MR-LoRA presents the most significant overhead. By utilizing a small subset of replay data to train the base model as a native router, MR-LoRA requires the storage of additional routing parameters alongside the task-specific weights, culminating in the highest storage overhead during inference among all evaluated methods.
\section{Conclusion}

\begin{wrapfigure}{r}{0.55\textwidth}
    \centering
    \includegraphics[clip, width=0.55\textwidth, height=0.6\textwidth]{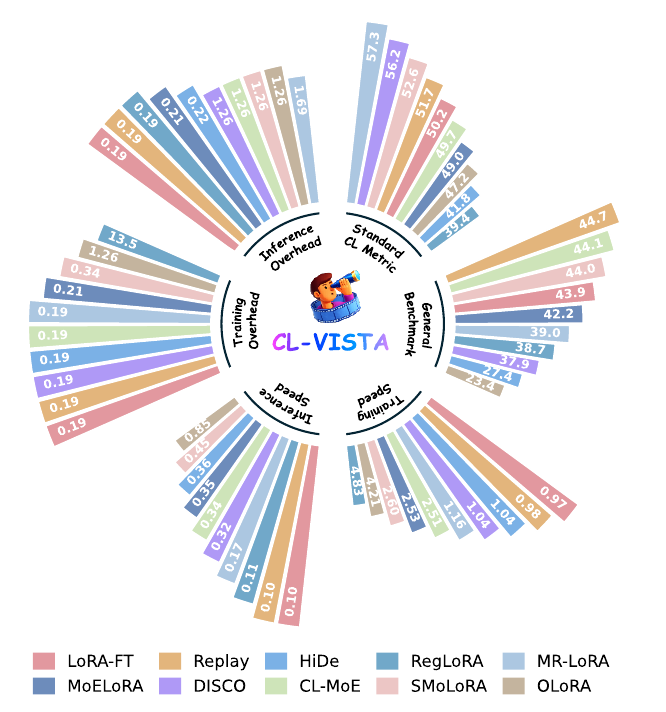}
    \vspace{-20pt}
    \caption{CL-VISTA Leaderboard.
    }
    \label{fig:leaderboard}
\end{wrapfigure}

In this paper, we propose CL-VISTA, the first continual video understanding benchmark tailored for Video-LLMs. To expose genuine catastrophic forgetting, CL-VISTA comprises 8 diverse tasks ranging from perception to reasoning. Furthermore, we establish a comprehensive evaluation framework with 6 dimensions covering performance, computational efficiency, and storage overhead. We systematically evaluate 10 mainstream multimodal continual learning methods across 2 prominent architectures, Video-LLaVA and VideoLLaMA2. As demonstrated in Fig. \ref{fig:leaderboard}, our extensive analysis reveals that no existing method achieves comprehensive superiority. Specifically, existing approaches inevitably face a critical trade-off: they either sacrifice foundational generalization for task-specific gains or incur prohibitive computational and memory overheads, rendering them impractical for real-world deployment. By open-sourcing CL-VISTA, we aim to provide a rigorous testbed to advance continual learning research in Video-LLMs and broader multimodal foundation models.

\clearpage

\newpage
\appendix

\section{Appendix A: CL-VISTA Curation and Statistical Analysis}
\label{appendix_A}
\setcounter{subsection}{0}
\renewcommand{\thesubsection}{A.\arabic{subsection}}

\setcounter{figure}{0} 
\renewcommand{\thefigure}{A.\arabic{figure}}

\subsection{Data statistics}

\begin{figure}[h]
  \centering
  \includegraphics[width=0.8\textwidth]{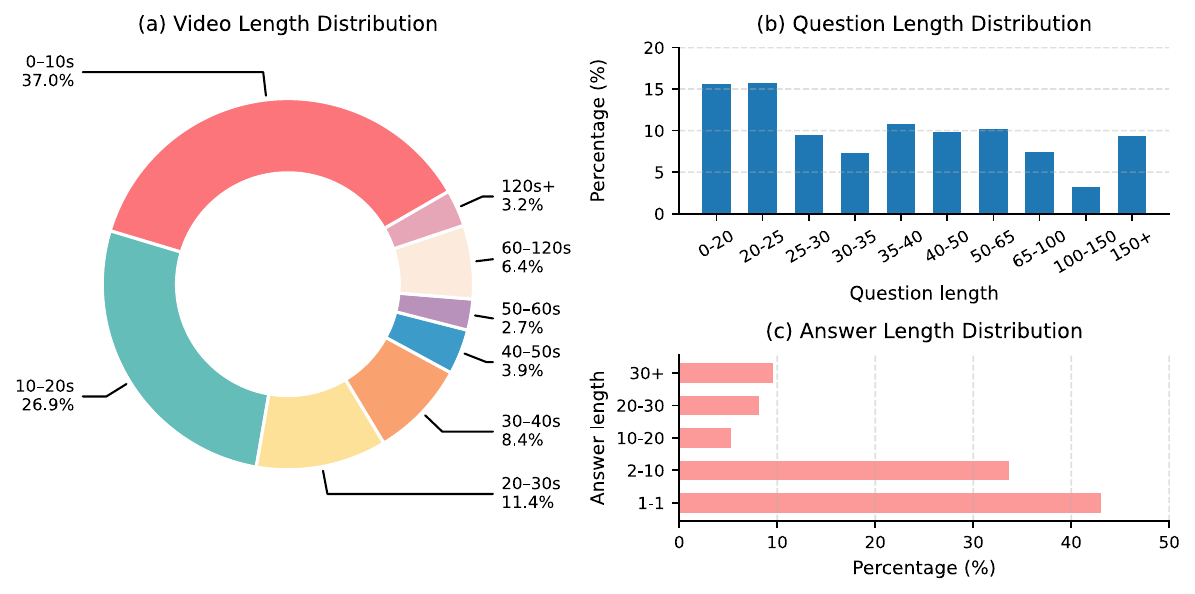}
  \vspace{-10pt}
  \caption{Data statistics of CL-VISTA.
  }
  \label{fig:app_data_statistics}
\end{figure}

As illustrated in Fig. \ref{fig:app_data_statistics}, CL-VISTA exhibits significant diversity across three key dimensions: video duration, question complexity, and answer granularity. (1) Video Length: The dataset covers a wide spectrum of temporal scales, ranging from short clips ($<$10s) to long-form videos exceeding 120s. A substantial portion (over 60\%) falls within the 0-20s range to capture dense event dynamics. (2) Question Length: The queries show a broad distribution, with a non-trivial presence of long-form instructions (over 150 words), reflecting the high complexity and multi-step reasoning required by our benchmark. (3) Answer Length: The responses span from concise single-word identifiers to detailed descriptive sentences (30+ words), providing a comprehensive testbed for both the recognition and generative capabilities of Video-LLMs.

\subsection{Model settings for data reconstruction}

We identify a significant scarcity of high-quality video-language data in specialized domains, particularly within Science and Sports. To bridge this gap, we implement a robust two-stage generation-discrimination pipeline to synthesize high-quality QA pairs from raw video data and associated metadata.

\textbf{Stage 1: High-quality QA Generation.} In the generation stage, we employ Gemini-2.5-Pro \cite{comanici2025gemini} as the primary generator $M_{gen}$. By integrating dense visual contexts with textual metadata, $M_{gen}$ synthesizes sentence-level QA pairs that are grounded in the temporal dynamics of the video.

\textbf{Stage 2: QA Quality Refinement.} To ensure the rigor of the benchmark, we first employ DeepSeek-V3.1 \cite{liu2024deepseek} as the blind test model $M_{blind}$. This model is tasked with answering questions without access to visual information; any QA pairs that can be correctly predicted through linguistic bias alone are pruned. Subsequently, we use GPT-5 \cite{singh2025openai} as the consistency judgment model $M_{consist}$ to verify the remaining candidates. Only QA pairs where both $M_{blind}$ and $M_{consist}$ achieve consensus on the ground truth are retained.

Task Enrichment. Finally, the filtered high-quality pairs undergo a format expansion phase using Gemini-2.5-Pro as the rewriting model $M_{rewrite}$. This process transforms concise sentence-level responses into a diverse set of task formats, including Multiple-choice, True/False, Short-answer, and Free-form descriptions, thereby ensuring the multifaceted nature of CL-VISTA.

\subsection{Visualization of each tasks}

\begin{figure}[h]
  \centering
  \begin{subfigure}{0.48\textwidth}
    \includegraphics[width=\textwidth]{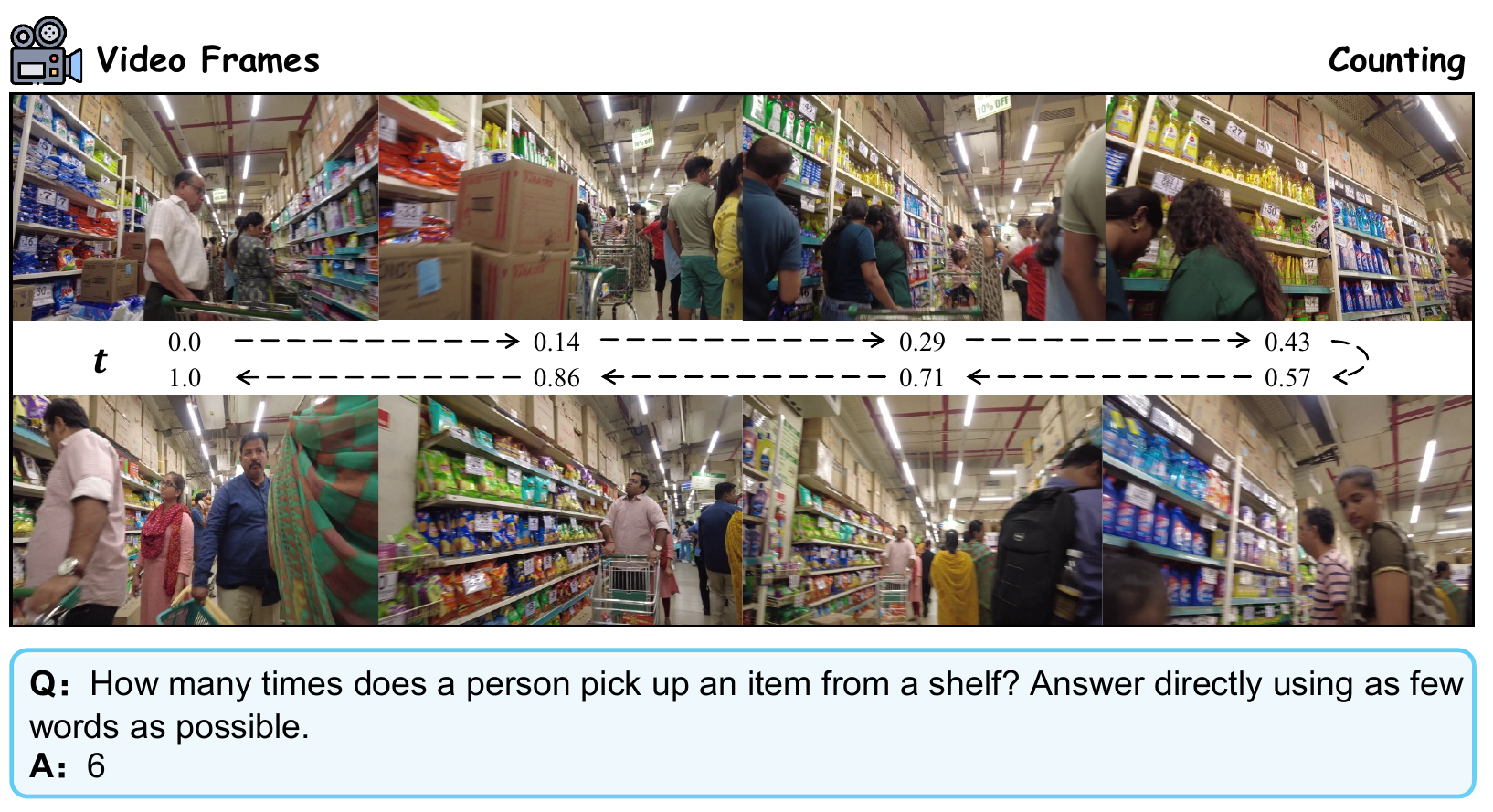}
  \end{subfigure}
  \hfill
  \begin{subfigure}{0.48\textwidth}
    \includegraphics[width=\textwidth]{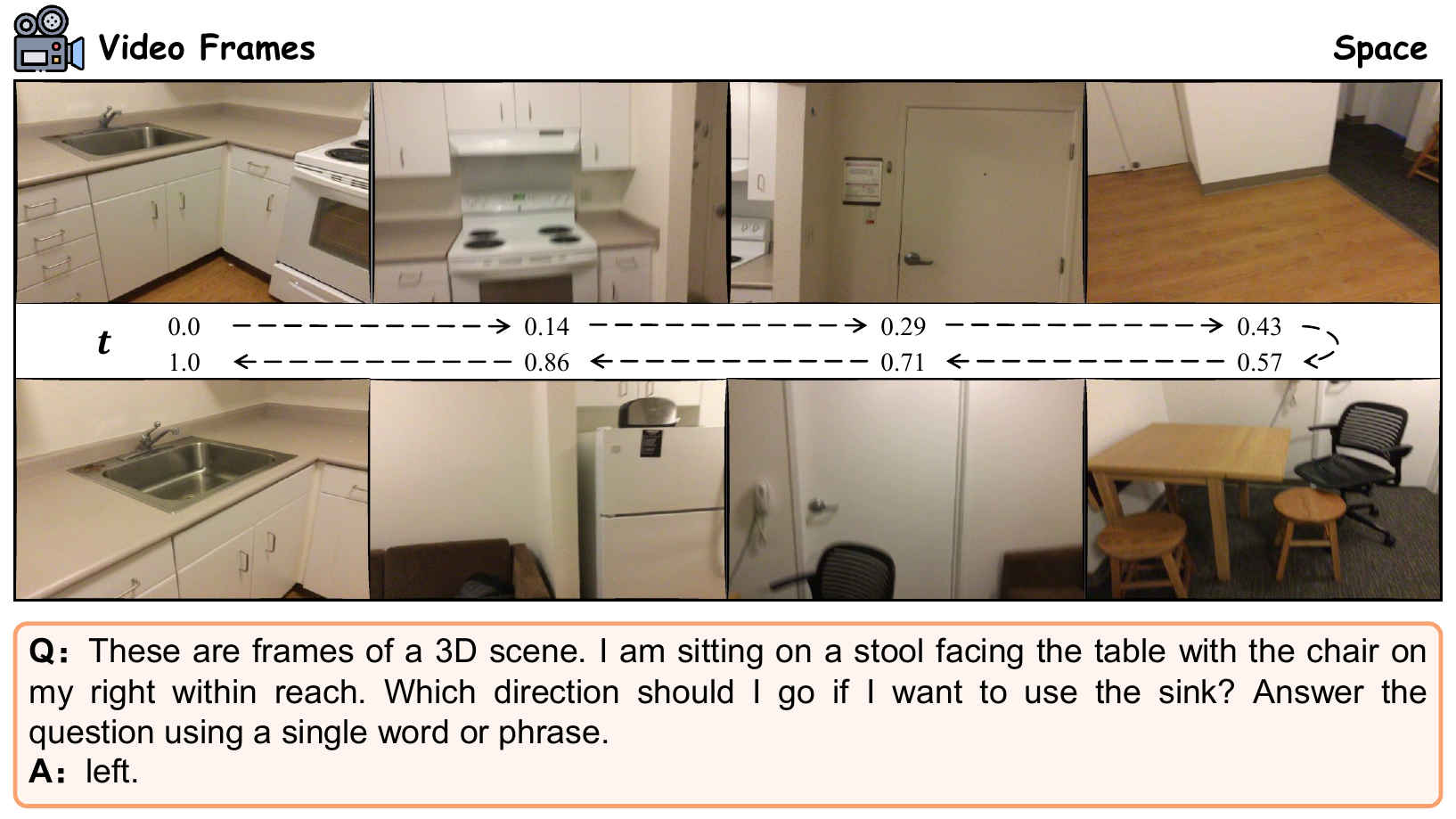}
  \end{subfigure}
  \caption{Visualizations of Counting and Space tasks.}
  \label{fig:app_visualization_1}
\end{figure}

\begin{figure}[h]
  \centering
  \begin{subfigure}{0.48\textwidth}
    \includegraphics[width=\textwidth]{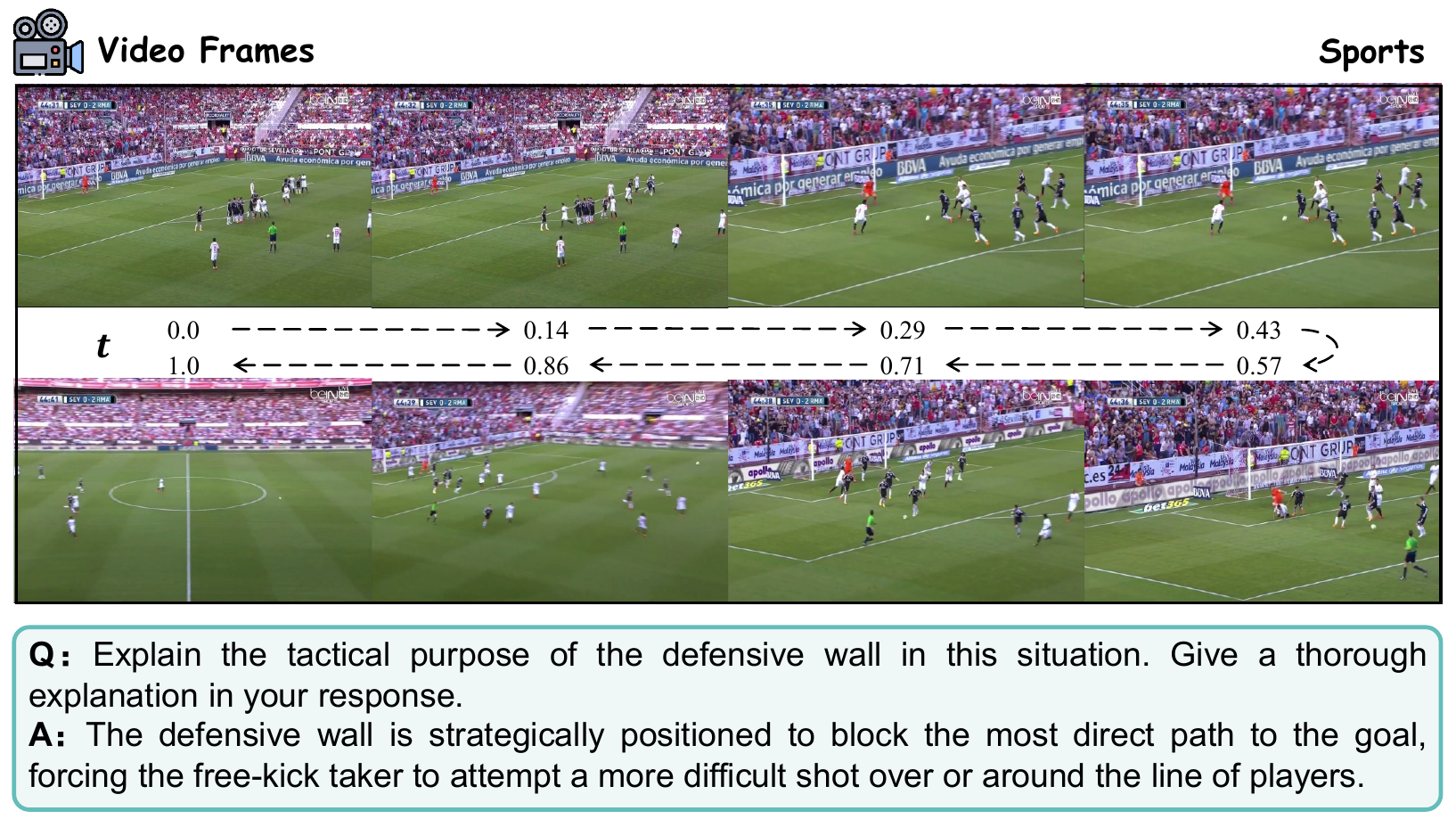}
  \end{subfigure}
  \hfill
  \begin{subfigure}{0.48\textwidth}
    \includegraphics[width=\textwidth]{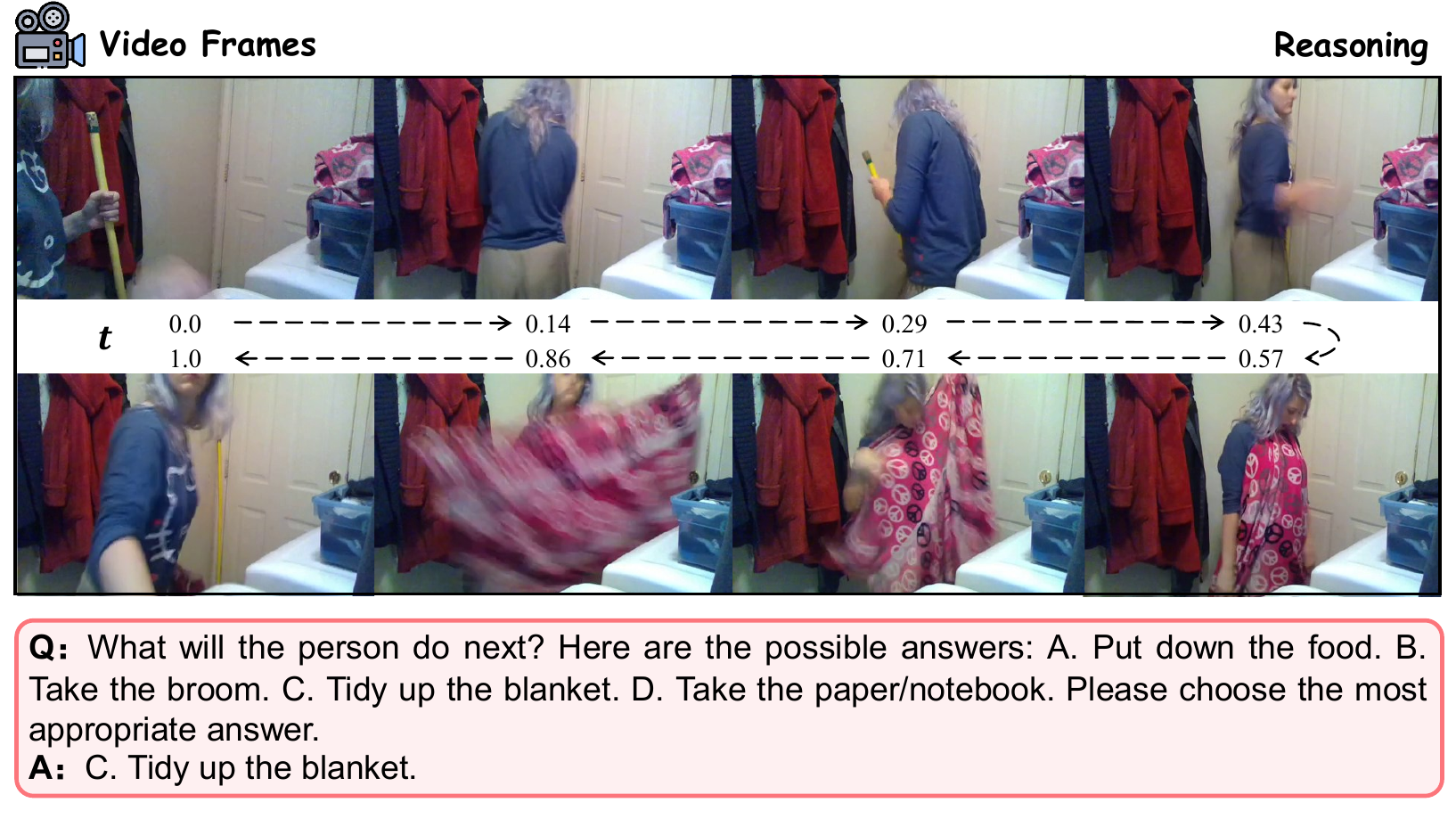}
  \end{subfigure}
  \caption{Visualizations of Sports and Reasoning tasks.}
  \label{fig:app_visualization_4}
\end{figure}

To provide a more intuitive understanding of the proposed benchmark, we present representative visual examples for each task category in CL-VISTA, as illustrated in Fig. \ref{fig:app_visualization_1} through Fig. \ref{fig:app_visualization_4}. These examples encompass a broad spectrum of real-world scenarios, including dynamic sports analysis, complex GUI operations, specialized scientific reasoning, and traffic monitoring. By covering diverse visual domains, CL-VISTA evaluates models across multiple dimensions, ranging from coarse-grained scene recognition to fine-grained temporal tracking and logical inference. Each visualization provides the source video frames, the corresponding long-form instruction, and the ground-truth response, highlighting the significant task heterogeneity and the video understanding and reasoning capabilities required to master the benchmark.

\begin{figure}[h]
  \centering
  \begin{subfigure}{0.48\textwidth}
    \includegraphics[width=\textwidth]{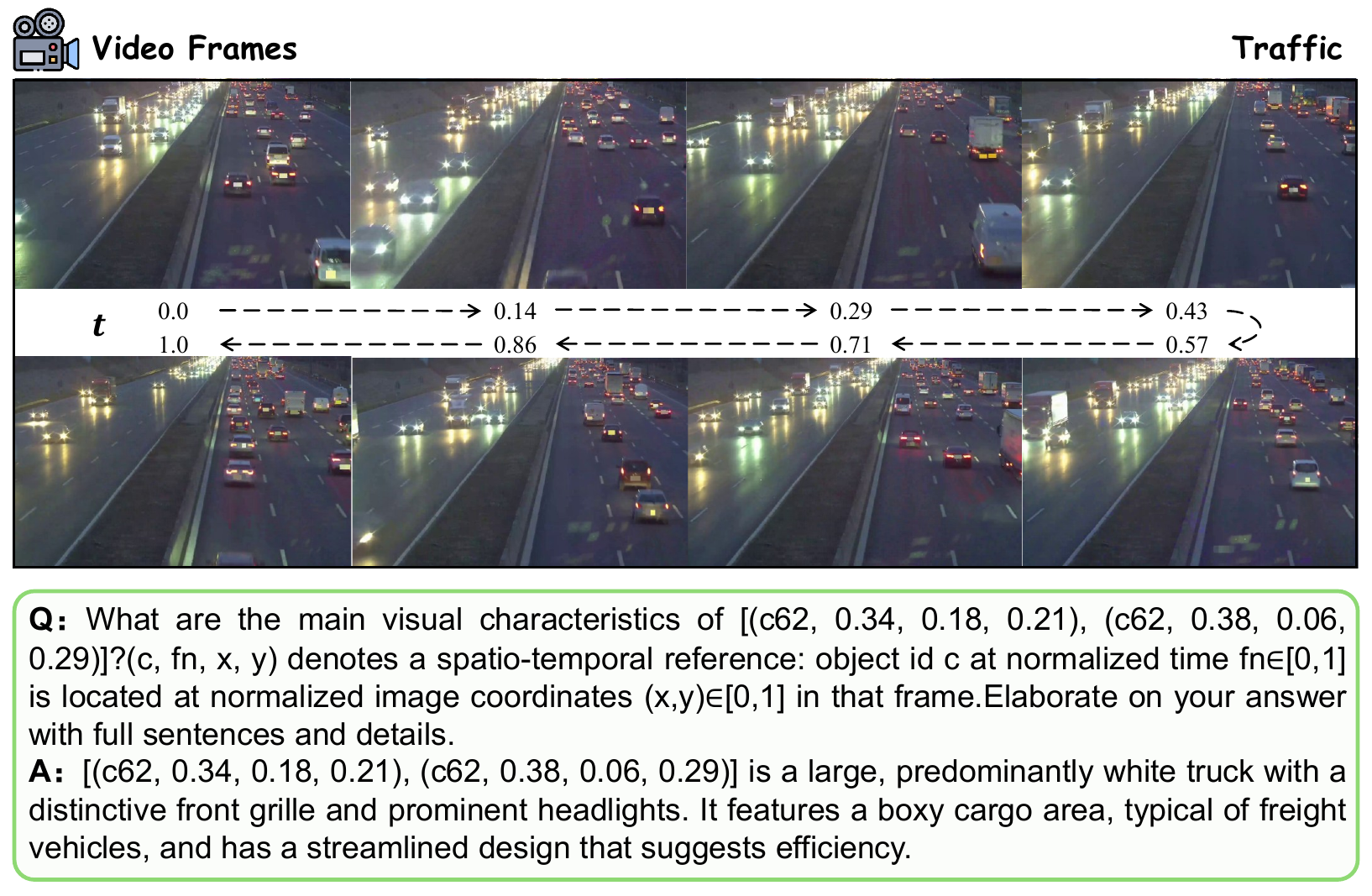}
  \end{subfigure}
  \hfill
  \begin{subfigure}{0.48\textwidth}
    \includegraphics[width=\textwidth]{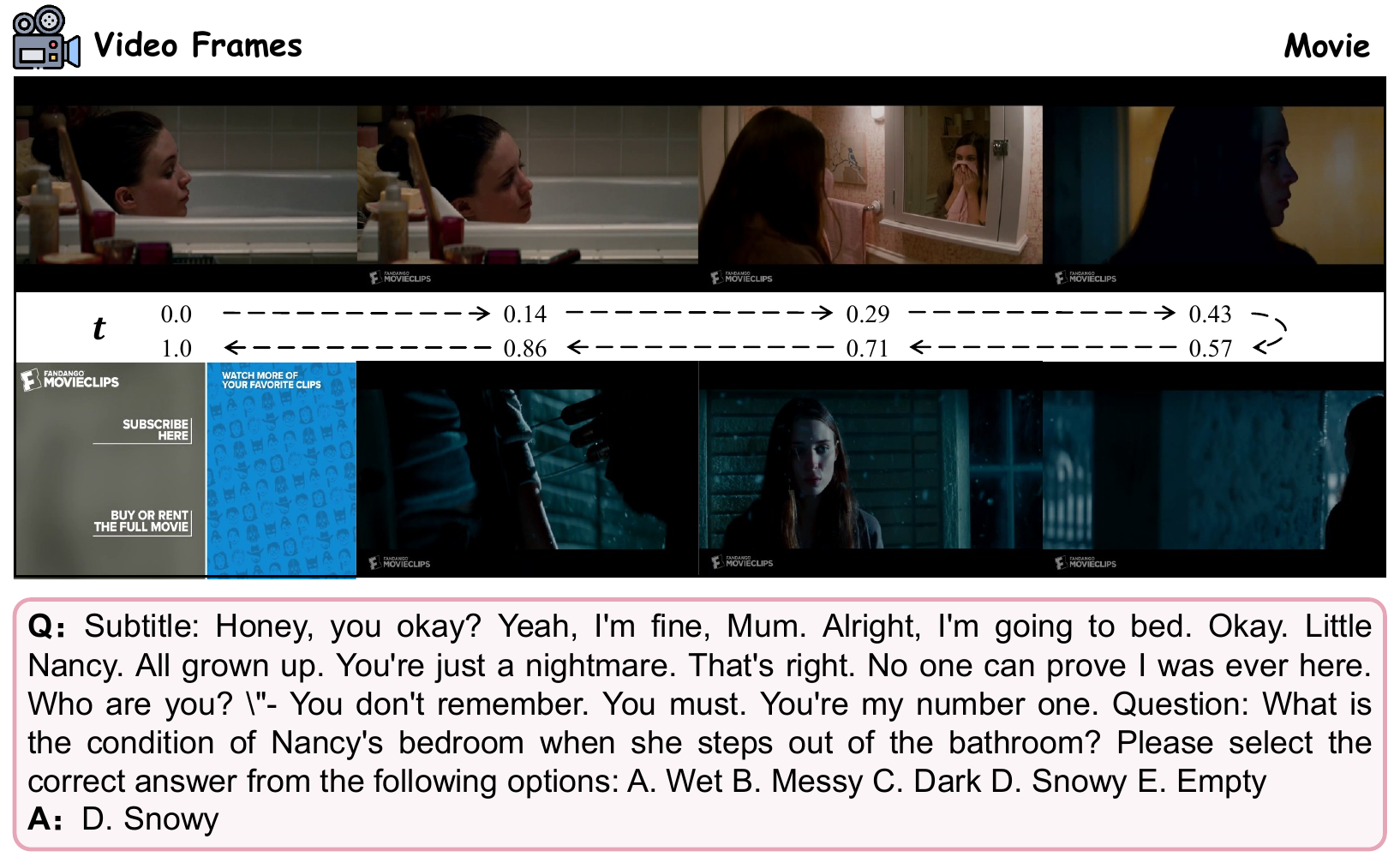}
  \end{subfigure}
  \caption{Visualizations of Traffic and Movie tasks.}
\end{figure}

\begin{figure}[h]
  \centering
  \begin{subfigure}{0.48\textwidth}
    \includegraphics[width=\textwidth]{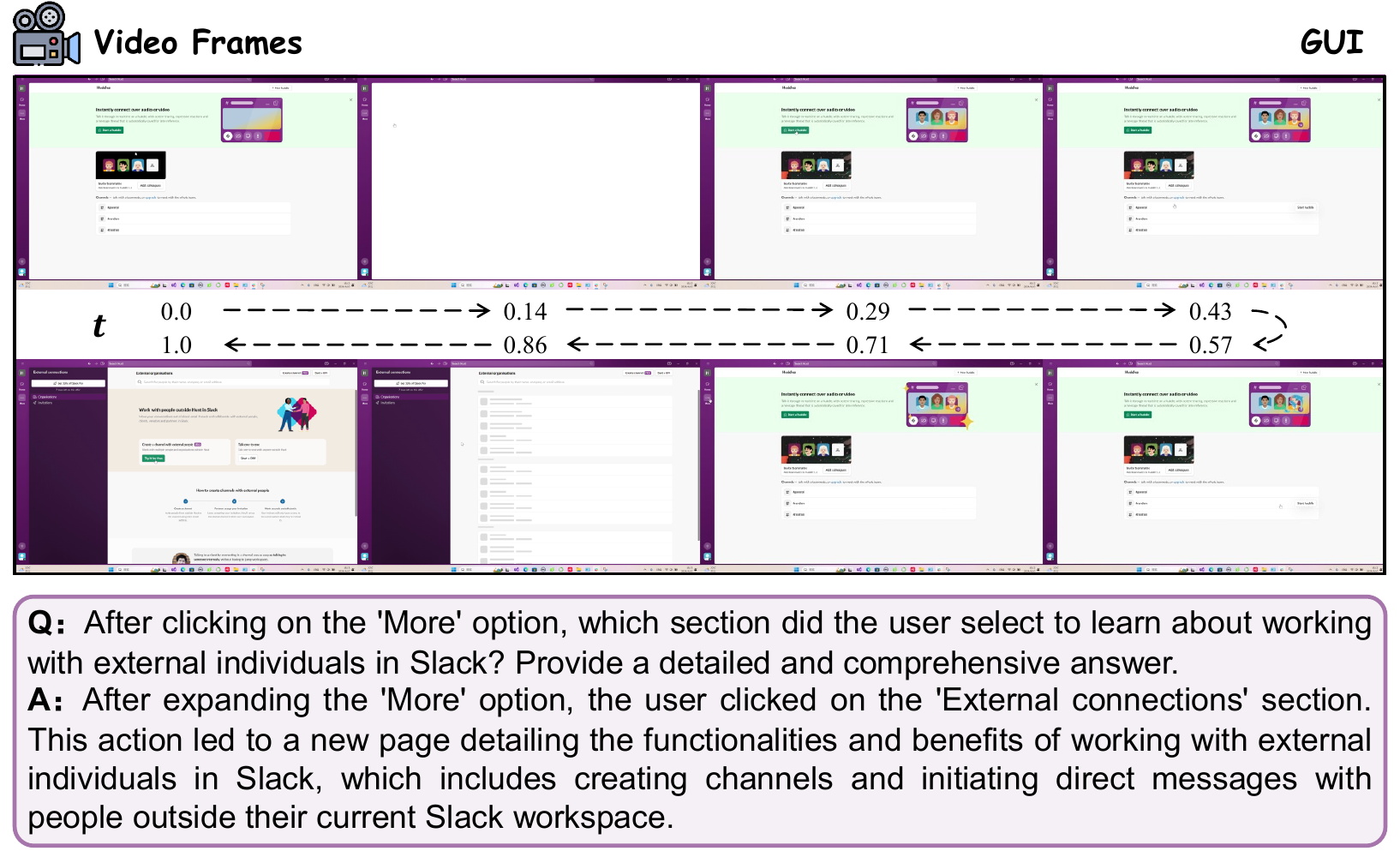}
  \end{subfigure}
  \hfill
  \begin{subfigure}{0.48\textwidth}
    \includegraphics[width=\textwidth]{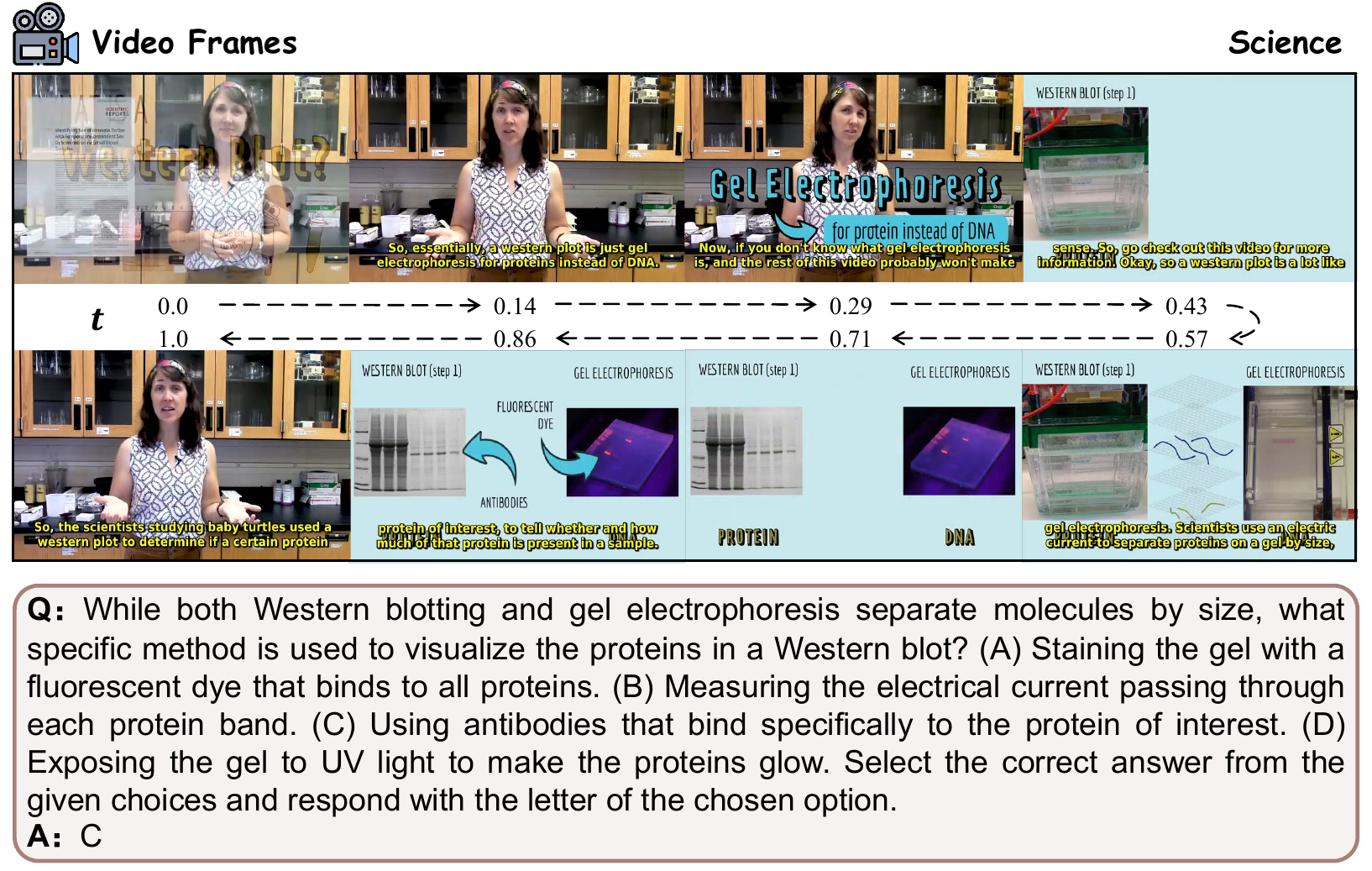}
  \end{subfigure}
  \caption{Visualizations of GUI and Science tasks.}
\end{figure}

\section{Appendix B: Details of Baseline Methods}
\label{appendix_B}
\setcounter{subsection}{0}
\renewcommand{\thesubsection}{B.\arabic{subsection}}

\setcounter{figure}{0} 
\renewcommand{\thefigure}{B.\arabic{figure}}

\setcounter{table}{0} 
\renewcommand{\thetable}{B.\arabic{table}}

In this section, we briefly describe each compared method; implementation details and hyperparameter settings are provided in Table \ref{tab:app_methods_detail}.

\begin{itemize}
    \item \textbf{LoRA-FT} is a straightforwa rd fine-tuning baseline that introduces trainable low-rank adaptation modules into the transformer blocks of the LLM while keeping the vision encoder frozen. Since no mechanism is employed to counteract catastrophic forgetting, it is typically regarded as the performance lower bound in continual learning benchmarks.
    \item \textbf{Replay} mitigates catastrophic forgetting by retaining a small subset of samples from each previous task and mixing them with the current task's training data at every subsequent stage. In our experiments, 100 samples per task are preserved in the replay buffer. By explicitly revisiting a limited amount of past data during training, this method partially alleviates forgetting and is commonly used as a practical upper-reference baseline in continual learning settings.
    \item \textbf{MR-LoRA} addresses catastrophic forgetting by allocating a dedicated task-specific LoRA module for each incoming task, ensuring strict parameter isolation across tasks. A small set of replay samples is used to additionally fine-tune the model itself as a task-expert router, which ensures routing accuracy but inevitably introduces substantial parameter overhead and inference latency.
    \item \textbf{OLoRA} mitigates catastrophic forgetting by incrementally learning each new task within a LoRA subspace that is constrained to be orthogonal to those of all previous tasks. Specifically, a dedicated set of LoRA parameters is introduced for each incoming task, and an orthogonality regularization term is added to the training objective to minimize interference with past task knowledge, while all previous LoRA parameters remain frozen.
    \item \textbf{RegLoRA} mitigates essential forgetting by identifying and protecting the parameters most critical to previously acquired knowledge. Upon completing each task, the top-magnitude elements in the LoRA weight update matrix are identified as key elements; when training on subsequent tasks, a regularization loss is applied to constrain these positions toward zero, preventing new LoRA updates from overwriting the parameters where prior knowledge is stored.
    \item \textbf{MoELoRA} mitigates catastrophic forgetting by replacing the single LoRA adapter with a pool of multiple identical yet independent LoRA experts inserted into the LLM. A trainable router models a probability distribution over the expert pool and dynamically combines their outputs for each input, enabling different experts to specialize in capturing distinct task knowledge. Since the shared expert pool is updated across all tasks without explicit isolation, knowledge interference across tasks can still occur as the number of tasks grows.
    \item \textbf{ModalPrompt} mitigates catastrophic forgetting through a prompt learning framework that requires no data replay or model expansion. A dedicated set of task-specific prototype prompts is learned for each incoming task in the joint vision-language feature space, and a prompt fusion module integrates knowledge from all previous frozen prototypes during training. In inference, a dual-modality guided selection module retrieves only the most relevant prompts for each input, preventing interference from irrelevant task knowledge while keeping computational complexity tractable.
    \item \textbf{HiDe} mitigates catastrophic forgetting through a hierarchical decoupling strategy motivated by CKA similarity analysis, which reveals that the top layer of the LLM focuses predominantly on task-specific knowledge while the remaining layers encode more generic, cross-task representations. Based on this observation, a dedicated task-specific LoRA is maintained and expanded at the top layer for each new task, with expert weights dynamically computed via dual-modality similarity matching against image-text anchors stored during training. For all other layers, the LoRA weights of all learned tasks are fused into a single set of parameters to consolidate shared general knowledge, avoiding the parameter explosion that plagues purely expansion-based approaches.
    \item \textbf{CL-MoE} mitigates catastrophic forgetting through two complementary components built upon a LoRA-based MoE framework. A Dual-Router MoE (RMoE) captures both instance-level and task-level expert assignments simultaneously, robustly allocating the most appropriate experts from local and global perspectives for each input. A Dynamic Momentum MoE (MMoE) then classifies the selected experts into task-shared and task-specific categories, and updates all expert parameters via a momentum strategy that balances retention of prior knowledge against absorption of new task knowledge.
    \item \textbf{DISCO} organizes task-specific knowledge into dedicated LoRA subspaces within a dynamic cache maintained on the global server, using an identity token matching mechanism to route and aggregate client uploads into their corresponding subspaces without privacy leakage. During inference, a Subspace Selective Activation module computes similarity scores between the test instruction's text embedding and all global identity tokens, applying softmax-normalized activation factors to amplify relevant subspaces while suppressing irrelevant ones.
    \item \textbf{SMoLoRA} addresses a dual form of catastrophic forgetting in continual visual instruction tuning, where MLLMs suffer degradation in both visual understanding and instruction following abilities as new tasks are learned. It introduces a separable routing strategy that independently selects the most suitable LoRA blocks for each domain, and an adaptive fusion module that performs a weighted combination of the routed outputs, enabling specialized adaptation while reducing cross-task interference.
\end{itemize}

\begin{table}[t]
  \caption{Training configurations and PEFT settings for all compared methods on CL-VISTA. Unless otherwise noted, all settings are kept identical across both base models, Video-LLaVA and VideoLLaMA2.}
  \label{tab:app_methods_detail}
  \centering
  \resizebox{0.8\textwidth}{!}{
  \setlength{\aboverulesep}{0pt}
  \setlength{\belowrulesep}{0pt}
  \renewcommand{\arraystretch}{1.2}
  \setlength{\tabcolsep}{5pt}
  \begin{tabular}{lccccc}
    \toprule
    \textbf{Method} & \textbf{LR} & \textbf{Projector LR} & \textbf{Epoch} & \textbf{PEFT} & \textbf{Parameter setting} \\
    \midrule
    LoRA-FT     & 1e-4 & 1e-5 & 1 & LoRA   & rank = 32, expert num = 1 \\
    Replay      & 1e-4 & 1e-5 & 1 & LoRA   & rank = 32, expert num = 1 \\
    MR-LoRA     & 1e-4 & 1e-5 & 1 & LoRA   & rank = 256, expert num = 8 \\
    O-LoRA      & 1e-4 & 1e-5 & 1 & LoRA   & rank = 256, expert num = 8 \\
    RegLoRA     & 1e-4 & 1e-5 & 1 & LoRA   & rank = 32, expert num = 1 \\
    MoELoRA     & 1e-4 & 1e-5 & 1 & LoRA   & rank = 32, expert num = 8 \\
    ModalPrompt & 1e-4 & 1e-5 & 1 & Prompt & prefix\_len = 10, expert num = 8 \\
    HiDe        & 1e-4 & 1e-5 & 1 & LoRA   & rank = 256, expert num = 8 \\
    CL-MoE      & 1e-4 & 1e-5 & 1 & LoRA   & rank = 256, expert num = 8 \\
    SMoLoRA     & 1e-4 & 1e-5 & 1 & LoRA   & rank = 256, expert num = 8 \\
    DISCO       & 1e-4 & 1e-5 & 1 & LoRA   & rank = 256, expert num = 8 \\
    \bottomrule
  \end{tabular}
  }
\end{table}

\section{Appendix C: Details of Evaluation Metrics}
\label{appendix_C}
\setcounter{subsection}{0}
\renewcommand{\thesubsection}{C.\arabic{subsection}}

\setcounter{figure}{0} 
\renewcommand{\thefigure}{C.\arabic{figure}}

\setcounter{table}{0} 
\renewcommand{\thetable}{C.\arabic{table}}

\subsection{Standard CL Evaluation Metrics}

Let $a_{i,j}$ denote the accuracy on task $j$ evaluated immediately after the model finishes training on task $i$, and let $T$ denote the total number of tasks. The four standard continual learning metrics are defined as follows.

\begin{itemize}
    \item \textbf{Mean Finetune Accuracy (MFT)}: The average accuracy on each task evaluated immediately after its training concludes, corresponding to the diagonal entries of the accuracy matrix. This metric quantifies the model's learning capability on new tasks and serves as an empirical upper bound on performance, assuming no catastrophic forgetting.
    \begin{equation}
        \text{MFT} = \frac{1}{T} \sum_{i=1}^{T} a_{i,i}
    \end{equation}

    \item \textbf{Mean Final Accuracy (MFN)}: The average accuracy across all tasks measured at the end of the entire training sequence, corresponding to the last row of the accuracy matrix. It reflects the overall knowledge retained by the model after learning all tasks.
    \begin{equation}
        \text{MFN} = \frac{1}{T} \sum_{j=1}^{T} a_{T,j}
    \end{equation}

    \item \textbf{Backward Transfer (BWT)}: Measures the influence of learning new tasks on the performance of previously learned tasks. It is computed as the average difference between the final accuracy of each task and the accuracy obtained immediately after its initial training. A negative BWT value is a direct indicator of catastrophic forgetting.
    \begin{equation}
        \text{BWT} = \frac{1}{T-1} \sum_{j=1}^{T-1} \left( a_{T,j} - a_{j,j} \right)
    \end{equation}

    \item \textbf{Mean Average Accuracy (MAA)}: Provides a holistic measure of performance throughout the entire learning process. After each task $i$ is trained, the average accuracy over all tasks seen so far is computed; MAA is then obtained by averaging these values across all training steps.
    \begin{equation}
        \text{MAA} = \frac{1}{T} \sum_{i=1}^{T} \frac{1}{i} \sum_{j=1}^{i} a_{i,j}
    \end{equation}
\end{itemize}

\subsection{LLM-as-Judge Evaluation Metric}

While the four standard CL metrics defined above characterize the overall continual learning dynamics, a critical question remains: how to determine the correctness of each individual prediction. Conventional rule-based evaluation methods rely on exact string matching or simple pattern extraction, which can introduce spurious forgetting artifacts. For instance, if the ground-truth answer is \texttt{2} but the model responds with \texttt{two}, a string-matching evaluator would incorrectly penalize this as a wrong answer, even though the model has correctly understood the question. Such false negatives conflate poor instruction-following with genuine knowledge forgetting, making it difficult to faithfully measure forgetting.

To mitigate this issue, we adopt an LLM-as-Judge scoring strategy, leveraging the powerful Qwen3-30A3B-Instruct model to evaluate each prediction on a scale from 0 to 5 based on semantic correctness rather than surface-level matching. This approach is robust to minor variations in answer format, synonymous expressions, and imperfect instruction following, thereby providing a more faithful reflection of the model's true knowledge retention. The detailed prompt used for scoring is shown in Figure~\ref{fig:LLM-as-Judge}.

\begin{figure}[h]
    \begin{tcolorbox}[
        title={LLM-based Evaluation Prompt},
        colback=white,
        colframe=gray!60,
        colbacktitle=gray!40,
        coltitle=black,
        fonttitle=\small\bfseries,
        boxrule=0.5pt,
        arc=3pt,
        left=6pt, right=6pt, top=4pt, bottom=4pt
    ]
    
    \small
    \textbf{[System Prompt]}
    
    \smallskip
    You are an intelligent chatbot designed for evaluating the correctness of generative outputs for question-answer pairs. Your task is to compare the predicted answer with the correct answer and determine the matching score.
    
    \smallskip
    \textbf{INSTRUCTIONS:}
    \begin{itemize}[leftmargin=1.5em, itemsep=0pt, topsep=2pt]
        \item Focus on the meaningful match between the predicted answer and the correct answer.
        \item Consider synonyms or paraphrases as valid matches.
        \item Evaluate the correctness of the prediction compared to the answer and provide a score from 0 to 5:
        \begin{itemize}[leftmargin=1.5em, itemsep=0pt, topsep=2pt]
            \item \textbf{0}: No match at all
            \item \textbf{1}: Very poor match
            \item \textbf{2}: Poor match
            \item \textbf{3}: Fair match
            \item \textbf{4}: Good match
            \item \textbf{5}: Excellent / Perfect match
        \end{itemize}
    \end{itemize}
    
    \noindent\rule{\linewidth}{0.4pt}
    
    \textbf{[User Prompt]}
    
    \smallskip
    Please evaluate the following video-based question-answer pair:
    
    \smallskip
    \textbf{Question:} \{Question\}\\
    \textbf{Correct Answer:} \{Answer\}\\
    \textbf{Predicted Answer:} \{Predicted Answer\}
    
    \smallskip
    Provide your evaluation only as a score where the score is an integer value between 0 and 5, with 5 indicating the highest meaningful match. Please generate the response in the form of a Python dictionary string with key \texttt{`score'} only, where the value is an INTEGER between 0 and 5. DO NOT PROVIDE ANY OTHER OUTPUT TEXT OR EXPLANATION. Only provide the Python dictionary string. For example: \texttt{\{‘score': 4\}}.
    
    \end{tcolorbox}
    \vspace{-10pt}
    \caption{Prompt template used for LLM-as-Judge evaluation.}
    \vspace{-15pt}
    \label{fig:LLM-as-Judge}
\end{figure}

\subsection{Details of General Video Benchmark}

To investigate whether the knowledge learned throughout the continual learning process can be transferred to unseen tasks, we conduct evaluations across a range of general video benchmarks, which are detailed below.

\begin{itemize}
    \item \textbf{NExT-QA} is a VideoQA benchmark designed to evaluate causal and temporal reasoning over video content, with a particular focus on complex object interactions in everyday activities. The dataset covers a diverse range of real-life scenarios, including family interactions, children at play, social gatherings, sports, pets, and musical performances. We evaluate the final checkpoint of each method on the NExT-QA validation set, which comprises 4,996 multiple-choice questions, using accuracy as the primary metric.
    \item \textbf{MMVU} is a comprehensive, expert-level, multi-disciplinary benchmark for evaluating foundation models on video understanding, spanning 27 subjects across four core disciplines: Science, Healthcare, Humanities \& Social Sciences, and Engineering. We evaluate each method on the MMVU validation set, which comprises 1,000 QA pairs in both multiple-choice and open-ended formats. We employ Qwen3-30B-A3B-Instruct as the judge model. Specifically, given a question, its ground-truth answer, and the model's response, the judge model is tasked with extracting the final answer from the model's output and assessing its correctness.
    \item \textbf{MVBench} is a comprehensive multimodal video benchmark with 20 tasks that require temporal reasoning beyond single-frame analysis, assessing MLLMs’ temporal perception in open-world settings. Due to missing source videos for some tasks, we evaluate on 19 tasks (excluding Pose), using 3,800 multiple-choice questions and reporting accuracy.
    \item \textbf{LongVideoBench} is a multiple-choice benchmark designed for long-context multimodal video understanding, with a particular emphasis on referred reasoning questions that require long frame sequences and cannot be adequately addressed by single or sparsely sampled frames. We evaluate each method on the validation set, which comprises 752 videos and 1,337 multiple-choice questions, using accuracy as the primary metric. All evaluations are conducted without the use of subtitles.
    \item \textbf{MMBench-Video} is a quantitative benchmark designed to rigorously evaluate the video understanding capabilities of MLLMs. It comprises long-form, topically diverse videos sourced from the web, accompanied by high-quality visual questions crafted by human volunteers across a broad range of fine-grained capability dimensions. Rather than adopting the official evaluation protocol that relies on GPT-4, we employ Qwen3-30B-A3B-Instruct as the judge model to assess the semantic alignment between each model's output and the ground-truth answer. Specifically, given a question, its ground-truth answer, and the model's response, the judge model scores the degree of meaningful alignment on a scale from 0 to 100, and the final performance metric is reported as the average of these scores.
\end{itemize}

\section{Appendix D: Detailed Results}
\label{appendix_D}
\setcounter{subsection}{0}
\renewcommand{\thesubsection}{D.\arabic{subsection}}

\setcounter{figure}{0} 
\renewcommand{\thefigure}{D.\arabic{figure}}

\setcounter{table}{0} 
\renewcommand{\thetable}{D.\arabic{table}}

\subsection{Results on VideoLLaMA2}

Tables~\ref{tab:videollama2_cl} and~\ref{tab:general_video_videollama2} report results on VideoLLaMA2, with findings consistent with those on Video-LLaVA. Parameter-expansion-based methods again demonstrate strong resistance to catastrophic forgetting: MR-LoRA achieves a BWT of $-0.03$ alongside the best MFN and MAA scores, while DISCO follows closely with a BWT of $-1.23$. However, both methods show consistent performance drops on general video understanding benchmarks relative to the LoRA-FT baseline, particularly on MMBench-Video. This highlights a persistent tension between anti-forgetting capability and knowledge transferability, where task-specific memory is well preserved but fails to generalize to unseen tasks.

\begin{table*}[h]
    \centering
    \caption{Performance comparison of different methods on VideoLLaMA2 using standard continual learning metrics. The best results are highlighted in \textbf{bold}, and the second-best results are \underline{underlined}.}
    \vspace{-5pt}
    \label{tab:videollama2_cl}
    \renewcommand{\arraystretch}{1.1} 

    \resizebox{0.9\linewidth}{!}{
    \begin{tabular}{@{} l *{8}{c} *{4}{c} @{}}
        \toprule
        Method & 
        \vhead{Count.} & \vhead{Space} & \vhead{Traffic} & \vhead{Movie} & 
        \vhead{GUI} & \vhead{Science} & \vhead{Sports} & \vhead{Reason.} & 
        \phead{MFT ($\uparrow$)} & \phead{MFN ($\uparrow$)} & \phead{MAA ($\uparrow$)} & \phead{BWT ($\uparrow$)} \\ 
        \midrule
        Zero-shot  & 45.82 & 48.02 & 46.40 & 49.69 & 68.05 & 65.58 & 47.72 & 63.62 & 54.36 & -- & -- & -- \\
        Individual & 57.86 & 64.22 & 77.11 & 85.13 & 80.11 & 86.79 & 91.04 & 84.74 & 78.38 & -- & -- & -- \\
        Joint & 58.32 & 67.44 & 74.65 & 85.54 & 80.24 & 85.27 & 91.22 & 85.40 & 78.51 & -- & -- & -- \\
        LoRA-FT    & 46.72 & 50.10 & 60.89 & 69.09 & 67.25 & 82.05 & 89.10 & 84.92 & 78.44 & 68.77 & 64.62 & -11.05 \\
        \midrule
        \rowcolor{myblue!10} \multicolumn{13}{c}{\textbf{Replay-based Methods}} \\
        \midrule
        Replay     & 53.65 & 56.29 & 66.24 & 75.14 & 74.13 & 84.03 & 87.12 & \textbf{86.53} & 77.22 & 72.89 & 65.94 & -4.94 \\
        MR-LoRA      & \textbf{56.54} & \underline{63.35} & \textbf{76.85} & \textbf{84.99} & \textbf{79.97} & \textbf{87.13} & \textbf{91.22} & \underline{85.75} & \textbf{78.25} & \textbf{78.23} & \textbf{69.40} & \underline{-0.03} \\
        \midrule
        \rowcolor{myblue!10} \multicolumn{13}{c}{\textbf{Regularization-based Methods}} \\
        \midrule
        OLoRA     & 50.31 & 56.44 & 51.96 & 56.24 & 85.72 & 76.49 & 85.91 & 64.26 & 65.64 & 65.92 & 59.17 & \textbf{0.32} \\
        RegLoRA   & 42.73 & 45.46 & 54.55 & 54.35 & 60.54 & 62.21 & 55.26 & 84.99 & 76.37 & 57.51 & 55.51 & -21.55 \\
        \midrule
        \rowcolor{myblue!10} \multicolumn{13}{c}{\textbf{Architecture-based Methods}} \\
        \midrule
        MoELoRA    & 49.12 & 48.71 & 58.81 & 71.23 & 67.28 & 81.73 & 87.29 & 85.11 & \underline{78.00} & 68.66 & 63.27 & -10.67 \\
        ModalPrompt& 46.05 & 44.22 & 48.05 & 51.25 & 61.88 & 54.67 & 41.65 & 81.19 & 64.46 & 53.62 & 48.68 & -12.39 \\
        CL-MoE     & 48.18 & 49.36 & 58.79 & 68.29 & 63.49 & 79.28 & 88.51 & 84.83 & 77.53 & 67.59 & 64.50 & -11.36 \\
        HiDe       & 41.38 & 39.26 & 35.62 & 52.17 & 57.18 & 48.69 & 57.78 & 58.59 & 64.07 & 48.83 & 56.90 & -17.41 \\
        SMoLoRA    & \underline{56.06} & 53.65 & 62.91 & 78.68 & 71.30 & 82.62 & 89.88 & 84.80 & 77.89 & 72.49 & 67.10 & -6.18 \\
        DISCO      & 56.03 & \textbf{66.54} & \underline{68.66} & \underline{84.87} & \underline{77.52} & \underline{84.49} & \underline{89.89} & 84.71 & 77.66 & \underline{76.59} & \underline{69.29} & -1.23 \\
        \bottomrule
    \end{tabular}}
\end{table*}

\begin{table*}[h]
    \centering
    \vspace{10pt}
    \caption{General video understanding performance of various CL methods on VideoLLaMA2. We highlight the relative change compared to the LoRA-FT baseline (\textcolor{upcolor}{green} for improvement, \textcolor{downcolor}{red} for decline).}
    \vspace{-5pt}
    \label{tab:general_video_videollama2}
    \renewcommand{\arraystretch}{1.1}
    \setlength{\tabcolsep}{5pt}

    \resizebox{0.8\linewidth}{!}{
    \begin{tabular}{@{} l ccccc c @{}}
        \toprule
        \textbf{Method} & \textbf{MMVU} & \textbf{MVBench} & \makecell[c]{\textbf{LongVideo}\\\textbf{Bench}} & 
        \makecell[c]{\textbf{MMBench-}\\\textbf{Video}} & 
        \textbf{NExTQA} & \textbf{Average} \\
        \midrule
        Zero-shot  & 34.70 & 52.63 & 44.43 & 35.92 & 38.37 & 41.21 \\
        Joint      & 40.50 & 51.61 & 43.98 & 38.95 & 60.31 & 47.07 \\
        LoRA-FT    & 36.30 & 50.58 & 44.28 & 35.97 & 55.52 & 44.53 \\
        \midrule
        \rowcolor{myblue!10} \multicolumn{7}{c}{\textbf{Replay-based Methods}} \\
        Replay     & 39.10 \up{2.80} & 52.84 \up{2.26} & 47.57 \up{3.29} & 38.91 \up{2.94} & 65.33 \up{9.81} & 48.75 \up{4.22} \\
        MR-LoRA    & 35.50 \down{0.80} & 47.00 \down{3.58} & 44.35 \up{0.07} & 37.93 \up{1.96} & 35.33 \down{20.19} & 40.02 \down{4.51} \\
        \midrule
        \rowcolor{myblue!10} \multicolumn{7}{c}{\textbf{Regularization-based Methods}} \\
        O-LoRA     & 38.30 \up{2.00} & 40.42 \down{10.16} & 34.70 \down{9.58} & 16.03 \down{19.94} & 27.32 \down{28.20} & 31.35 \down{13.18} \\
        RegLoRA       & 29.00 \down{7.30} & 49.92 \down{0.66} & 38.07 \down{6.21} & 32.38 \down{3.59} & 40.03 \down{15.49} & 37.88 \down{6.65} \\
        \midrule
        \rowcolor{myblue!10} \multicolumn{7}{c}{\textbf{Architecture-based Methods}} \\
        MoELoRA    & 37.00 \up{0.70} & 52.26 \up{1.68} & 45.03 \up{0.75} & 35.27 \down{0.70} & 60.77 \up{5.25} & 46.07 \up{1.54} \\
        HiDE       & 18.60 \down{17.70} & 37.39 \down{13.19} & 30.44 \down{13.84} & 23.31 \down{12.66} & 28.48 \down{27.04} & 27.64 \down{16.89} \\
        CL-MoE     & 32.60 \down{3.70} & 49.39 \down{1.19} & 42.18 \down{2.10} & 32.70 \down{3.27} & 55.04 \down{0.48} & 42.38 \down{2.15} \\
        DISCO      & 36.40 \up{0.10} & 52.26 \up{1.68} & 44.88 \up{0.60} & 37.12 \up{1.15} & 42.05 \down{13.47} & 42.54 \down{1.99} \\
        SMOLoRA    & 36.70 \up{0.40} & 49.24 \down{1.34} & 44.65 \up{0.37} & 35.38 \down{0.59} & 59.79 \up{4.27} & 45.15 \up{0.62} \\
        ModalPrompt & 27.80 \down{8.50} & 48.18 \down{2.40} & 40.16 \down{4.12} & 30.32 \down{5.65} & 37.77 \down{17.75} & 36.85 \down{7.68} \\
        \bottomrule
    \end{tabular}}
\end{table*}

\subsection{Continual Learning Result Matrices}

We report the continual learning result matrices for each method on Video-LLaVA and VideoLLaMA2 in Tables~\ref{tab:cl_lora_ft}--\ref{tab:cl_modalprompt}.

\begin{table}[t]
\centering
\caption{Continual learning results of \textbf{LoRA-FT} on different Video-LLM backbones.}
\label{tab:cl_lora_ft}
\begin{subtable}{0.48\textwidth}
\centering
\resizebox{\textwidth}{!}{%
\begin{tabular}{lcccccccc}
\toprule
\textbf{Video-LLaVA} & Count. & Space & Traffic & Movie & GUI & Science & Sports & Reason. \\
\midrule
Counting & 59.69 & & & & & & & \\
Space    & 48.71 & 71.19 & & & & & & \\
Traffic  & 54.01 & 59.67 & 68.35 & & & & & \\
Movie    & 51.21 & 62.01 & 64.59 & 88.80 & & & & \\
GUI      & 50.44 & 55.70 & 60.80 & 85.10 & 78.48 & & & \\
Science  & 33.13 & 49.34 & 52.76 & 80.12 & 72.37 & 83.68 & & \\
Sports   & 32.59 & 51.01 & 51.75 & 75.32 & 71.32 & 81.79 & 89.75 & \\
Star     & 41.13 & 54.81 & 55.84 & 74.61 & 68.80 & 78.20 & 88.90 & 86.86 \\
\bottomrule
\end{tabular}%
}
\end{subtable}
\hfill
\begin{subtable}{0.48\textwidth}
\centering
\resizebox{\textwidth}{!}{%
\begin{tabular}{lcccccccc}
\toprule
\textbf{VideoLLaMA2} & Count. & Space & Traffic & Movie & GUI & Science & Sports & Reason. \\
\midrule
Counting & 57.86 & & & & & & & \\
Space    & 57.30 & 63.42 & & & & & & \\
Traffic  & 56.76 & 58.85 & 77.79 & & & & & \\
Movie    & 54.90 & 58.11 & 72.12 & 85.93 & & & & \\
GUI      & 46.55 & 56.42 & 68.66 & 83.18 & 79.71 & & & \\
Science  & 40.62 & 51.36 & 59.43 & 78.21 & 74.90 & 86.73 & & \\
Sports   & 30.82 & 51.69 & 56.65 & 70.70 & 74.33 & 84.36 & 91.12 & \\
Star     & 46.72 & 50.10 & 60.89 & 69.09 & 67.25 & 82.05 & 89.10 & 84.92 \\
\bottomrule
\end{tabular}%
}
\end{subtable}

\end{table}

\begin{table}[t]
\centering
\caption{Continual learning results of \textbf{MoELoRA} on different Video-LLM backbones.}
\begin{subtable}{0.48\textwidth}
\centering
\resizebox{\textwidth}{!}{%
\begin{tabular}{lcccccccc}
\toprule
\textbf{Video-LLaVA} & Count. & Space & Traffic & Movie & GUI & Science & Sports & Reason. \\
\midrule
Counting & 58.30 & & & & & & & \\
Space    & 52.26 & 67.97 & & & & & & \\
Traffic  & 56.04 & 57.89 & 67.75 & & & & & \\
Movie    & 55.49 & 58.93 & 64.84 & 84.74 & & & & \\
GUI      & 54.63 & 54.94 & 57.98 & 82.02 & 77.38 & & & \\
Science  & 37.04 & 46.69 & 51.29 & 78.62 & 70.61 & 81.93 & & \\
Sports   & 35.95 & 52.11 & 48.71 & 73.89 & 70.44 & 80.61 & 88.60 & \\
Star     & 40.44 & 52.88 & 54.27 & 72.46 & 67.27 & 79.29 & 87.47 & 84.23 \\
\bottomrule
\end{tabular}%
}
\end{subtable}
\hfill
\begin{subtable}{0.48\textwidth}
\centering
\resizebox{\textwidth}{!}{%
\begin{tabular}{lcccccccc}
\toprule
\textbf{VideoLLaMA2} & Count. & Space & Traffic & Movie & GUI & Science & Sports & Reason. \\
\midrule
Counting & 57.46 & & & & & & & \\
Space    & 54.46 & 63.17 & & & & & & \\
Traffic  & 49.76 & 56.71 & 76.46 & & & & & \\
Movie    & 54.35 & 53.99 & 69.73 & 85.85 & & & & \\
GUI      & 41.53 & 51.67 & 65.40 & 83.92 & 79.94 & & & \\
Science  & 40.84 & 47.71 & 57.03 & 80.47 & 73.51 & 85.87 & & \\
Sports   & 29.30 & 47.97 & 55.83 & 76.68 & 74.15 & 84.58 & 90.12 & \\
Star     & 49.12 & 48.71 & 58.81 & 71.23 & 67.28 & 81.73 & 87.29 & 85.11 \\
\bottomrule
\end{tabular}%
}
\end{subtable}

\end{table}

\begin{table}[t]
\centering
\caption{Continual learning results of \textbf{Replay} on different Video-LLM backbones.}
\begin{subtable}{0.48\textwidth}
\centering
\resizebox{\textwidth}{!}{%
\begin{tabular}{lcccccccc}
\toprule
\textbf{Video-LLaVA} & Count. & Space & Traffic & Movie & GUI & Science & Sports & Reason. \\
\midrule
Counting & 59.69 & & & & & & & \\
Space    & 56.50 & 69.45 & & & & & & \\
Traffic  & 56.93 & 61.73 & 69.01 & & & & & \\
Movie    & 56.38 & 60.52 & 58.57 & 87.62 & & & & \\
GUI      & 55.06 & 54.63 & 64.45 & 84.48 & 78.27 & & & \\
Science  & 51.00 & 54.84 & 55.32 & 81.40 & 73.79 & 83.18 & & \\
Sports   & 55.09 & 55.27 & 54.75 & 80.70 & 74.69 & 81.54 & 89.96 & \\
Star     & 53.39 & 55.13 & 59.27 & 74.78 & 70.91 & 79.67 & 86.70 & 87.76 \\
\bottomrule
\end{tabular}%
}
\end{subtable}
\hfill
\begin{subtable}{0.48\textwidth}
\centering
\resizebox{\textwidth}{!}{%
\begin{tabular}{lcccccccc}
\toprule
\textbf{VideoLLaMA2} & Count. & Space & Traffic & Movie & GUI & Science & Sports & Reason. \\
\midrule
Counting & 58.41 & & & & & & & \\
Space    & 58.24 & 60.09 & & & & & & \\
Traffic  & 54.93 & 57.38 & 76.53 & & & & & \\
Movie    & 53.27 & 53.97 & 68.71 & 86.93 & & & & \\
GUI      & 54.47 & 53.45 & 70.52 & 85.64 & 79.85 & & & \\
Science  & 51.14 & 52.35 & 62.06 & 81.93 & 75.66 & 87.04 & & \\
Sports   & 52.19 & 54.21 & 61.65 & 79.59 & 76.53 & 86.03 & 91.07 & \\
Star     & 52.42 & 53.61 & 67.43 & 73.60 & 73.48 & 83.53 & 86.41 & 84.74 \\
\bottomrule
\end{tabular}%
}
\end{subtable}

\end{table}

\begin{table}[t]
\centering
\caption{Continual learning results of \textbf{DISCO} on different Video-LLM backbones.}
\begin{subtable}{0.48\textwidth}
\centering
\resizebox{\textwidth}{!}{%
\begin{tabular}{lcccccccc}
\toprule
\textbf{Video-LLaVA} & Count. & Space & Traffic & Movie & GUI & Science & Sports & Reason. \\
\midrule
Counting & 58.99 & & & & & & & \\
Space    & 59.68 & 72.50 & & & & & & \\
Traffic  & 59.29 & 72.78 & 67.19 & & & & & \\
Movie    & 59.46 & 72.67 & 66.70 & 90.56 & & & & \\
GUI      & 59.38 & 71.84 & 62.48 & 90.76 & 78.65 & & & \\
Science  & 59.34 & 72.75 & 60.42 & 90.38 & 78.07 & 78.30 & & \\
Sports   & 59.27 & 72.13 & 60.48 & 90.26 & 78.44 & 78.76 & 88.71 & \\
Star     & 59.69 & 72.58 & 61.18 & 90.45 & 78.27 & 78.61 & 88.16 & 90.04 \\
\bottomrule
\end{tabular}%
}
\end{subtable}
\hfill
\begin{subtable}{0.48\textwidth}
\centering
\resizebox{\textwidth}{!}{%
\begin{tabular}{lcccccccc}
\toprule
\textbf{VideoLLaMA2} & Count. & Space & Traffic & Movie & GUI & Science & Sports & Reason. \\
\midrule
Counting & 57.39 & & & & & & & \\
Space    & 56.74 & 64.90 & & & & & & \\
Traffic  & 57.75 & 68.64 & 74.73 & & & & & \\
Movie    & 56.24 & 69.22 & 72.85 & 85.15 & & & & \\
GUI      & 57.38 & 68.38 & 72.50 & 84.85 & 79.06 & & & \\
Science  & 56.35 & 67.92 & 68.84 & 84.76 & 78.21 & 85.18 & & \\
Sports   & 56.75 & 67.39 & 67.77 & 85.18 & 77.67 & 84.27 & 90.17 & \\
Star     & 56.03 & 66.54 & 68.66 & 84.87 & 77.52 & 84.49 & 89.89 & 84.71 \\
\bottomrule
\end{tabular}%
}
\end{subtable}

\end{table}

\begin{table}[t]
\centering
\caption{Continual learning results of \textbf{HiDe} on different Video-LLM backbones.}
\begin{subtable}{0.48\textwidth}
\centering
\resizebox{\textwidth}{!}{%
\begin{tabular}{lcccccccc}
\toprule
\textbf{Video-LLaVA} & Count. & Space & Traffic & Movie & GUI & Science & Sports & Reason. \\
\midrule
Counting & 59.90 & & & & & & & \\
Space    & 58.17 & 69.86 & & & & & & \\
Traffic  & 58.46 & 66.05 & 58.46 & & & & & \\
Movie    & 57.83 & 63.31 & 57.00 & 81.57 & & & & \\
GUI      & 55.16 & 56.22 & 48.61 & 78.72 & 69.70 & & & \\
Science  & 50.99 & 51.98 & 44.91 & 72.35 & 65.73 & 72.20 & & \\
Sports   & 38.21 & 47.28 & 40.06 & 65.86 & 62.10 & 65.47 & 69.95 & \\
Star     & 41.35 & 47.32 & 39.01 & 64.46 & 60.29 & 64.36 & 67.46 & 51.26 \\
\bottomrule
\end{tabular}%
}
\end{subtable}
\hfill
\begin{subtable}{0.48\textwidth}
\centering
\resizebox{\textwidth}{!}{%
\begin{tabular}{lcccccccc}
\toprule
\textbf{VideoLLaMA2} & Count. & Space & Traffic & Movie & GUI & Science & Sports & Reason. \\
\midrule
Counting & 56.68 & & & & & & & \\
Space    & 57.86 & 64.08 & & & & & & \\
Traffic  & 58.37 & 58.57 & 59.37 & & & & & \\
Movie    & 59.60 & 53.85 & 53.83 & 74.67 & & & & \\
GUI      & 56.27 & 54.02 & 50.26 & 74.57 & 67.23 & & & \\
Science  & 51.39 & 49.32 & 45.51 & 69.60 & 65.76 & 69.50 & & \\
Sports   & 40.74 & 45.98 & 33.07 & 55.31 & 59.26 & 56.64 & 62.42 & \\
Star     & 41.38 & 39.26 & 35.62 & 52.17 & 57.18 & 48.69 & 57.78 & 58.59 \\
\bottomrule
\end{tabular}%
}
\end{subtable}

\end{table}

\begin{table}[t]
\centering
\caption{Continual learning results of \textbf{CL-MoE} on different Video-LLM backbones.}
\begin{subtable}{0.48\textwidth}
\centering
\resizebox{\textwidth}{!}{%
\begin{tabular}{lcccccccc}
\toprule
\textbf{Video-LLaVA} & Count. & Space & Traffic & Movie & GUI & Science & Sports & Reason. \\
\midrule
Counting & 59.32 & & & & & & & \\
Space    & 51.97 & 69.15 & & & & & & \\
Traffic  & 55.14 & 59.74 & 66.50 & & & & & \\
Movie    & 51.91 & 66.59 & 61.97 & 86.77 & & & & \\
GUI      & 53.07 & 56.04 & 58.47 & 83.50 & 78.72 & & & \\
Science  & 34.14 & 48.40 & 51.82 & 79.53 & 72.21 & 82.81 & & \\
Sports   & 33.04 & 52.88 & 49.38 & 73.27 & 70.86 & 81.42 & 89.71 & \\
Star     & 42.79 & 55.77 & 54.15 & 74.28 & 67.65 & 78.06 & 87.79 & 86.58 \\
\bottomrule
\end{tabular}%
}
\end{subtable}
\hfill
\begin{subtable}{0.48\textwidth}
\centering
\resizebox{\textwidth}{!}{%
\begin{tabular}{lcccccccc}
\toprule
\textbf{VideoLLaMA2} & Count. & Space & Traffic & Movie & GUI & Science & Sports & Reason. \\
\midrule
Counting & 55.27 & & & & & & & \\
Space    & 55.83 & 61.92 & & & & & & \\
Traffic  & 54.22 & 64.29 & 73.94 & & & & & \\
Movie    & 55.47 & 59.53 & 70.62 & 86.24 & & & & \\
GUI      & 52.08 & 55.57 & 67.39 & 84.64 & 80.20 & & & \\
Science  & 47.68 & 50.83 & 59.74 & 80.24 & 74.50 & 86.65 & & \\
Sports   & 37.34 & 53.03 & 56.42 & 76.33 & 74.25 & 84.17 & 91.18 & \\
Star     & 48.18 & 49.36 & 58.79 & 68.29 & 63.49 & 79.28 & 88.51 & 84.83 \\
\bottomrule
\end{tabular}%
}
\end{subtable}

\end{table}

\begin{table}[t]
\centering
\caption{Continual learning results of \textbf{RegLoRA} on different Video-LLM backbones.}
\begin{subtable}{0.48\textwidth}
\centering
\resizebox{\textwidth}{!}{%
\begin{tabular}{lcccccccc}
\toprule
\textbf{Video-LLaVA} & Count. & Space & Traffic & Movie & GUI & Science & Sports & Reason. \\
\midrule
Counting & 59.01 & & & & & & & \\
Space    & 40.20 & 70.21 & & & & & & \\
Traffic  & 44.24 & 39.09 & 68.68 & & & & & \\
Movie    & 34.63 & 39.74 & 45.89 & 87.27 & & & & \\
GUI      & 38.49 & 43.85 & 50.15 & 36.58 & 75.05 & & & \\
Science  & 26.94 & 48.63 & 44.16 & 29.91 & 58.94 & 76.71 & & \\
Sports   & 40.24 & 41.51 & 46.28 & 35.15 & 66.90 & 68.79 & 86.16 & \\
Star     & 45.20 & 43.47 & 46.89 & 37.23 & 59.17 & 60.36 & 52.92 & 86.51 \\
\bottomrule
\end{tabular}%
}
\end{subtable}
\hfill
\begin{subtable}{0.48\textwidth}
\centering
\resizebox{\textwidth}{!}{%
\begin{tabular}{lcccccccc}
\toprule
\textbf{VideoLLaMA2} & Count. & Space & Traffic & Movie & GUI & Science & Sports & Reason. \\
\midrule
Counting & 56.88 & & & & & & & \\
Space    & 36.62 & 62.70 & & & & & & \\
Traffic  & 43.08 & 45.44 & 74.64 & & & & & \\
Movie    & 41.71 & 47.44 & 48.12 & 84.69 & & & & \\
GUI      & 38.77 & 46.99 & 52.06 & 57.05 & 77.77 & & & \\
Science  & 40.39 & 50.30 & 48.99 & 50.07 & 62.68 & 82.99 & & \\
Sports   & 42.53 & 44.83 & 48.63 & 53.53 & 71.10 & 71.30 & 86.31 & \\
Star     & 42.73 & 45.46 & 54.55 & 54.35 & 60.54 & 62.21 & 55.26 & 84.99 \\
\bottomrule
\end{tabular}%
}
\end{subtable}

\end{table}

\begin{table}[t]
\centering
\caption{Continual learning results of \textbf{SMoLoRA} on different Video-LLM backbones.}
\begin{subtable}{0.48\textwidth}
\centering
\resizebox{\textwidth}{!}{%
\begin{tabular}{lcccccccc}
\toprule
\textbf{Video-LLaVA} & Count. & Space & Traffic & Movie & GUI & Science & Sports & Reason. \\
\midrule
Counting & 59.51 & & & & & & & \\
Space    & 58.31 & 71.50 & & & & & & \\
Traffic  & 57.10 & 66.53 & 68.65 & & & & & \\
Movie    & 56.55 & 67.08 & 68.26 & 88.21 & & & & \\
GUI      & 57.42 & 59.62 & 65.41 & 86.49 & 78.19 & & & \\
Science  & 51.23 & 56.89 & 56.32 & 84.22 & 76.55 & 83.47 & & \\
Sports   & 52.81 & 56.12 & 55.84 & 79.82 & 73.86 & 82.64 & 88.95 & \\
Star     & 51.66 & 59.45 & 57.70 & 80.64 & 71.17 & 81.87 & 87.31 & 86.92 \\
\bottomrule
\end{tabular}%
}
\end{subtable}
\hfill
\begin{subtable}{0.48\textwidth}
\centering
\resizebox{\textwidth}{!}{%
\begin{tabular}{lcccccccc}
\toprule
\textbf{VideoLLaMA2} & Count. & Space & Traffic & Movie & GUI & Science & Sports & Reason. \\
\midrule
Counting & 56.64 & & & & & & & \\
Space    & 55.83 & 67.92 & & & & & & \\
Traffic  & 57.53 & 68.11 & 73.31 & & & & & \\
Movie    & 57.52 & 61.59 & 71.44 & 85.08 & & & & \\
GUI      & 56.76 & 58.30 & 70.43 & 84.23 & 79.58 & & & \\
Science  & 52.97 & 54.37 & 61.39 & 82.12 & 76.93 & 84.99 & & \\
Sports   & 55.23 & 54.01 & 60.02 & 81.00 & 77.16 & 84.93 & 90.81 & \\
Star     & 56.06 & 53.65 & 62.91 & 78.68 & 71.30 & 82.62 & 89.88 & 84.80 \\
\bottomrule
\end{tabular}%
}
\end{subtable}

\end{table}

\begin{table}[t]
\centering
\caption{Continual learning results of \textbf{MR-LoRA} on different Video-LLM backbones.}
\begin{subtable}{0.48\textwidth}
\centering
\resizebox{\textwidth}{!}{%
\begin{tabular}{lcccccccc}
\toprule
\textbf{Video-LLaVA} & Count. & Space & Traffic & Movie & GUI & Science & Sports & Reason. \\
\midrule
Counting & 59.77 & & & & & & & \\
Space    & 59.82 & 72.50 & & & & & & \\
Traffic  & 59.79 & 72.54 & 67.51 & & & & & \\
Movie    & 59.77 & 72.50 & 67.97 & 89.87 & & & & \\
GUI      & 59.77 & 72.61 & 67.96 & 89.81 & 79.42 & & & \\
Science  & 59.86 & 72.50 & 67.54 & 89.90 & 79.44 & 83.24 & & \\
Sports   & 59.79 & 72.50 & 67.77 & 89.85 & 79.38 & 83.03 & 89.43 & \\
Star     & 59.72 & 72.50 & 67.99 & 89.85 & 79.46 & 82.89 & 89.46 & 89.75 \\
\bottomrule
\end{tabular}%
}
\end{subtable}
\hfill
\begin{subtable}{0.48\textwidth}
\centering
\resizebox{\textwidth}{!}{%
\begin{tabular}{lcccccccc}
\toprule
\textbf{VideoLLaMA2} & Count. & Space & Traffic & Movie & GUI & Science & Sports & Reason. \\
\midrule
Counting & 56.49 & & & & & & & \\
Space    & 56.49 & 63.35 & & & & & & \\
Traffic  & 56.49 & 63.35 & 76.94 & & & & & \\
Movie    & 57.01 & 63.35 & 76.60 & 85.00 & & & & \\
GUI      & 56.54 & 63.35 & 77.36 & 84.97 & 80.09 & & & \\
Science  & 56.48 & 63.35 & 77.00 & 85.00 & 80.04 & 87.15 & & \\
Sports   & 56.34 & 63.35 & 77.22 & 85.00 & 80.02 & 87.09 & 91.22 & \\
Star     & 56.54 & 63.35 & 76.85 & 84.99 & 79.97 & 87.13 & 91.22 & 85.75 \\
\bottomrule
\end{tabular}%
}
\end{subtable}

\end{table}

\begin{table}[t]
\centering
\caption{Continual learning results of \textbf{OLoRA} on different Video-LLM backbones.}
\begin{subtable}{0.48\textwidth}
\centering
\resizebox{\textwidth}{!}{%
\begin{tabular}{lcccccccc}
\toprule
\textbf{Video-LLaVA} & Count. & Space & Traffic & Movie & GUI & Science & Sports & Reason. \\
\midrule
Counting & 59.45 & & & & & & & \\
Space    & 57.84 & 67.15 & & & & & & \\
Traffic  & 58.13 & 66.54 & 59.71 & & & & & \\
Movie    & 55.83 & 66.96 & 55.59 & 73.44 & & & & \\
GUI      & 54.52 & 60.98 & 50.70 & 67.98 & 62.07 & & & \\
Science  & 46.97 & 57.31 & 43.34 & 64.14 & 60.45 & 66.35 & & \\
Sports   & 40.57 & 51.77 & 41.14 & 58.42 & 57.56 & 61.96 & 67.42 & \\
Star     & 41.05 & 52.53 & 41.00 & 58.82 & 57.94 & 62.37 & 67.46 & 46.97 \\
\bottomrule
\end{tabular}%
}
\end{subtable}
\hfill
\begin{subtable}{0.48\textwidth}
\centering
\resizebox{\textwidth}{!}{%
\begin{tabular}{lcccccccc}
\toprule
\textbf{VideoLLaMA2} & Count. & Space & Traffic & Movie & GUI & Science & Sports & Reason. \\
\midrule
Counting & 57.98 & & & & & & & \\
Space    & 56.11 & 63.70 & & & & & & \\
Traffic  & 54.61 & 56.65 & 55.71 & & & & & \\
Movie    & 54.56 & 52.34 & 52.47 & 74.30 & & & & \\
GUI      & 53.30 & 52.40 & 50.54 & 69.68 & 66.21 & & & \\
Science  & 49.41 & 47.55 & 43.86 & 66.28 & 67.52 & 72.52 & & \\
Sports   & 48.05 & 48.15 & 43.87 & 61.30 & 70.96 & 71.87 & 70.43 & \\
Star     & 50.31 & 56.44 & 51.96 & 56.24 & 85.72 & 76.49 & 85.91 & 64.26 \\
\bottomrule
\end{tabular}%
}
\end{subtable}

\end{table}

\begin{table}[t]
\centering
\caption{Continual learning results of \textbf{ModalPrompt} on different Video-LLM backbones.}
\label{tab:cl_modalprompt}
\begin{subtable}{0.48\textwidth}
\centering
\resizebox{\textwidth}{!}{%
\begin{tabular}{lcccccccc}
\toprule
\textbf{Video-LLaVA} & Count. & Space & Traffic & Movie & GUI & Science & Sports & Reason. \\
\midrule
Counting & 31.32 & & & & & & & \\
Space    & 31.85 & 26.13 & & & & & & \\
Traffic  & 32.48 & 15.64 & 58.64 & & & & & \\
Movie    & 31.24 & 26.41 & 41.62 & 38.93 & & & & \\
GUI      & 31.15 & 27.63 & 41.83 & 37.90 & 40.85 & & & \\
Science  & 32.09 & 26.34 & 41.70 & 38.33 & 40.62 & 56.18 & & \\
Sports   & 31.48 & 28.52 & 41.18 & 37.91 & 40.92 & 56.35 & 29.35 & \\
Star     & 31.81 & 26.32 & 41.38 & 38.14 & 41.55 & 56.20 & 29.20 & 36.76 \\
\bottomrule
\end{tabular}%
}
\end{subtable}
\hfill
\begin{subtable}{0.48\textwidth}
\centering
\resizebox{\textwidth}{!}{%
\begin{tabular}{lcccccccc}
\toprule
\textbf{VideoLLaMA2} & Count. & Space & Traffic & Movie & GUI & Science & Sports & Reason. \\
\midrule
Counting & 50.07 & & & & & & & \\
Space    & 41.11 & 61.57 & & & & & & \\
Traffic  & 40.89 & 49.88 & 64.53 & & & & & \\
Movie    & 35.00 & 43.81 & 46.33 & 73.39 & & & & \\
GUI      & 20.28 & 34.53 & 20.25 &  0.00 & 22.00 & & & \\
Science  & 42.07 & 48.55 & 49.61 & 50.72 & 68.99 & 81.90 & & \\
Sports   & 36.34 & 45.22 & 49.56 & 48.27 & 66.89 & 69.35 & 81.02 & \\
Star     & 46.05 & 44.22 & 48.05 & 51.25 & 61.88 & 54.67 & 41.65 & 81.19 \\
\bottomrule
\end{tabular}%
}
\end{subtable}

\end{table}

\clearpage
\newpage
{
    \small
    \bibliographystyle{ieeenat_fullname}
    \bibliography{main}
}

\end{document}